\documentstyle[examples,newapa,latex-acl,ipamacs]{article}
\title{AUTOMATED TONE TRANSCRIPTION}
\author{Steven Bird \\
        University of Edinburgh,
        Centre for Cognitive Science \\
        2 Buccleuch Place,
        Edinburgh, EH8 9LW, UK \\
        Internet: {\tt Steven.Bird@ed.ac.uk}
}

\font\ipatwelverm=wsuipa12

\font\ipanine=wsuipa9
\def\normalipa{\def\ipa{\ipatwelverm}}
\def\smallipa{\def\ipa{\ipanine}}
\normalipa
\def\trans#1{{\large #1}}
\def\transx#1{{\smallipa #1\normalipa}}

\def\cpp{C$\!${\raisebox{.3ex}{\scriptsize ++}}}
\def\ling#1{{\sl #1}}
\def\emph#1{{\em #1}}
\def\downstep{\mbox{$\downarrow$}}
\def\ds{\mbox{$\downarrow$}}
\def\upstep{\mbox{$\uparrow$}}
\def\us{\mbox{$\uparrow$}}
\def\mydot{\large .}
\def\hh{\makebox(0,0){\tiny $\circ$}}
\def\ll{\makebox(0,0){\footnotesize $\bullet$}}
\def\hl{\makebox(0,0){\footnotesize $\triangleleft$}}
\def\ldh{\makebox(0,0){\footnotesize $\triangleright$}}
\def\hhlab{$\circ$}
\def\lllab{$\bullet$}
\def\hllab{$\triangleleft$}
\def\ldhlab{$\triangleright$}

\def\tDH{\small \downstep H}
\def\tH{\small H}
\def\tL{\small L}

\def\MO{\smallipa\small m\openo\normalipa}
\def\SENG{\smallipa\small s\schwa\eng\normalipa}
\def\MBO{\smallipa\small mb\openo\normalipa}
\def\pair#1#2#3{
  \put(#1,63){\makebox(0,0)[t]{\strut #2}}
  \put(#1,50){\makebox(0,0)[t]{\strut #3}}
}
\def\negsp{\hspace{-0.2ex}}
\def\tbar#1{\bar{t}_{#1}}
\def\tpbar#1{\bar{t}'_{#1}}
\def\sbar#1{\bar{s}_{#1}}
\def\Rm#1{R_{#1}}
\def\R#1{R_{\mbox{\scriptsize #1}}}
\def\Pm#1{{\cal P}_{#1}}

\def\P#1{{\cal P}_{\mbox{\scriptsize #1}}}
\def\Pp#1{{\cal P}_{\mbox{\scriptsize #1}}'}

\def\Hdash{\raisebox{.8ex}{--}}
\def\Mdash{--}
\def\Ldash{\raisebox{-.8ex}{--}}

\begin{document}
\maketitle
\vspace{-0.5in}
\begin{abstract}
In this paper I report on an investigation into the problem of assigning
tones to pitch contours.  The proposed model is intended to serve as a
tool for phonologists working on instrumentally obtained pitch
data from tone languages.  Motivation and exemplification for the model
is provided by data taken from my fieldwork on Bamileke Dschang (Cameroon).
Following recent work by Liberman and others, I provide a parametrised
F$_0$ prediction function $\cal P$ which generates F$_0$ values from a
tone sequence, and I explore the asymptotic behaviour of downstep.
Next, I observe that transcribing a sequence $X$ of pitch
(i.e.~F$_0$) values
amounts to finding a tone sequence $T$ such that ${\cal P}(T) \approx X$.
This is a combinatorial optimisation problem, for which two
non-deterministic search techniques are provided:
a genetic algorithm and a simulated annealing algorithm.
Finally, two implementations---one for each technique---are described
and then compared using both
artificial and real data for sequences of up to 20 tones.
These programs can be adapted to other tone languages by adjusting
the F$_0$ prediction function.
\end{abstract}

\section{INTRODUCTION}

The wealth of literature on tone and intonation has amply demonstrated
that voice pitch (F$_0$) in speech is under independent linguistic
control.  In English, voice pitch alone can signal the distinction
between a statement and a question.  Similarly, in many tone
languages,voice pitch alone signals the tense of a verb.  Phonologists
usually describe a pitch contour much as they describe speech more
generally, namely as a sequence of discrete units (i.e.~a
transcription).  This is illustrated in Figure 1, where L indicates
a low tone and \downstep H indicates a downstepped high tone.
\begin{figure*}
\begin{center}
\setlength{\unitlength}{0.275mm}
\begin{picture}(560,240)(-20,40)
\put(0,70){\framebox(550,180){}}
\put(0,80){\line(1,0){5}}
\put(0,90){\line(1,0){5}}
\put(0,100){\line(1,0){10}} \put(-5,100){\makebox(0,0)[r]{\small 100}}
\put(0,110){\line(1,0){5}}
\put(0,120){\line(1,0){5}}
\put(0,130){\line(1,0){5}}
\put(0,140){\line(1,0){5}}
\put(0,150){\line(1,0){10}} \put(-5,150){\makebox(0,0)[r]{\small 150}}
\put(0,160){\line(1,0){5}}
\put(0,170){\line(1,0){5}}
\put(0,180){\line(1,0){5}}
\put(0,190){\line(1,0){5}}
\put(0,200){\line(1,0){10}} \put(-5,200){\makebox(0,0)[r]{\small 200}}
\put(0,210){\line(1,0){5}}
\put(0,220){\line(1,0){5}}
\put(0,230){\line(1,0){5}}
\put(0,240){\line(1,0){5}}
\put(-5,250){\makebox(0,0)[r]{\small Hz}}

\pair{40}{\tDH}{\MO}
\pair{65}{\tL}{\MBO}
\pair{90}{\tDH}{\MO}
\pair{120}{\tL}{\MBO}
\pair{145}{\tDH}{\MO}
\pair{175}{\tL}{\MBO}
\pair{200}{\tDH}{\MO}
\pair{230}{\tL}{\MBO}
\pair{255}{\tDH}{\MO}
\pair{285}{\tL}{\MBO}
\pair{310}{\tDH}{\MO}
\pair{335}{\tL}{\MBO}
\pair{360}{\tDH}{\MO}
\pair{385}{\tL}{\MBO}
\pair{415}{\tDH}{\MO}
\pair{440}{\tL}{\MBO}
\pair{465}{\tDH}{\MO}
\pair{495}{\tL}{\MBO}

\put(15,158){\mydot}	
\put(16,162){\mydot}	
\put(17,164){\mydot}	
\put(18,166){\mydot}	
\put(19,168){\mydot}	
\put(20,170){\mydot}	
\put(21,172){\mydot}	
\put(22,175){\mydot}	
\put(23,177){\mydot}	
\put(24,179){\mydot}	
\put(25,182){\mydot}	
\put(26,186){\mydot}	
\put(27,200){\mydot}	
\put(28,201){\mydot}	
\put(29,209){\mydot}	
\put(30,212){\mydot}	
\put(31,215){\mydot}	
\put(32,218){\mydot}	
\put(33,218){\mydot}	
\put(34,219){\mydot}	
\put(35,218){\mydot}	
\put(36,217){\mydot}	
\put(37,216){\mydot}	
\put(38,214){\mydot}	
\put(39,211){\mydot}	
\put(40,207){\mydot}	
\put(41,202){\mydot}	
\put(42,199){\mydot}	
\put(43,194){\mydot}	
\put(44,190){\mydot}	
\put(45,188){\mydot}	
\put(46,187){\mydot}	
\put(47,187){\mydot}	
\put(48,185){\mydot}	
\put(49,182){\mydot}	
\put(50,183){\mydot}	
\put(51,187){\mydot}	
\put(52,180){\mydot}	
\put(53,178){\mydot}	
\put(54,176){\mydot}	
\put(55,173){\mydot}	
\put(56,170){\mydot}	
\put(57,170){\mydot}	
\put(58,172){\mydot}	
\put(59,173){\mydot}	
\put(60,172){\mydot}	
\put(61,170){\mydot}	
\put(62,168){\mydot}	
\put(63,168){\mydot}	
\put(64,168){\mydot}	
\put(65,168){\mydot}	
\put(66,168){\mydot}	
\put(67,168){\mydot}	
\put(68,168){\mydot}	
\put(69,168){\mydot}	
\put(70,168){\mydot}	
\put(71,168){\mydot}	
\put(72,167){\mydot}	
\put(73,167){\mydot}	
\put(74,166){\mydot}	
\put(75,163){\mydot}	
\put(76,164){\mydot}	
\put(77,163){\mydot}	
\put(78,159){\mydot}	
\put(79,162){\mydot}	
\put(80,163){\mydot}	
\put(81,164){\mydot}	
\put(82,166){\mydot}	
\put(83,170){\mydot}	
\put(84,173){\mydot}	
\put(85,176){\mydot}	
\put(86,178){\mydot}	
\put(87,180){\mydot}	
\put(88,182){\mydot}	
\put(89,183){\mydot}	
\put(90,183){\mydot}	
\put(91,183){\mydot}	
\put(92,182){\mydot}	
\put(93,181){\mydot}	
\put(94,179){\mydot}	
\put(95,178){\mydot}	
\put(96,176){\mydot}	
\put(97,174){\mydot}	
\put(98,174){\mydot}	
\put(99,173){\mydot}	
\put(100,169){\mydot}	
\put(101,168){\mydot}	
\put(102,166){\mydot}	
\put(103,165){\mydot}	
\put(104,160){\mydot}	
\put(105,160){\mydot}	
\put(106,163){\mydot}	
\put(107,163){\mydot}	
\put(108,157){\mydot}	
\put(109,154){\mydot}	
\put(110,154){\mydot}	
\put(111,154){\mydot}	
\put(112,155){\mydot}	
\put(113,153){\mydot}	
\put(114,155){\mydot}	
\put(115,152){\mydot}	
\put(116,154){\mydot}	
\put(117,154){\mydot}	
\put(118,152){\mydot}	
\put(119,149){\mydot}	
\put(120,150){\mydot}	
\put(121,149){\mydot}	
\put(122,148){\mydot}	
\put(123,148){\mydot}	
\put(124,148){\mydot}	
\put(125,148){\mydot}	
\put(126,147){\mydot}	
\put(127,147){\mydot}	
\put(128,147){\mydot}	
\put(129,147){\mydot}	
\put(130,147){\mydot}	
\put(131,146){\mydot}	
\put(132,146){\mydot}	
\put(133,148){\mydot}	
\put(134,146){\mydot}	
\put(135,147){\mydot}	
\put(136,148){\mydot}	
\put(137,150){\mydot}	
\put(138,152){\mydot}	
\put(139,154){\mydot}	
\put(140,155){\mydot}	
\put(141,156){\mydot}	
\put(142,156){\mydot}	
\put(143,157){\mydot}	
\put(144,158){\mydot}	
\put(145,160){\mydot}	
\put(146,159){\mydot}	
\put(147,159){\mydot}	
\put(148,159){\mydot}	
\put(149,158){\mydot}	
\put(150,157){\mydot}	
\put(151,156){\mydot}	
\put(152,155){\mydot}	
\put(153,154){\mydot}	
\put(154,152){\mydot}	
\put(155,150){\mydot}	
\put(156,146){\mydot}	
\put(157,143){\mydot}	
\put(158,141){\mydot}	
\put(159,141){\mydot}	
\put(160,139){\mydot}	
\put(161,136){\mydot}	
\put(162,132){\mydot}	
\put(163,138){\mydot}	
\put(164,136){\mydot}	
\put(165,137){\mydot}	
\put(166,136){\mydot}	
\put(167,135){\mydot}	
\put(168,135){\mydot}	
\put(169,134){\mydot}	
\put(170,133){\mydot}	
\put(171,131){\mydot}	
\put(172,132){\mydot}	
\put(173,142){\mydot}	
\put(174,134){\mydot}	
\put(175,134){\mydot}	
\put(176,134){\mydot}	
\put(177,134){\mydot}	
\put(178,135){\mydot}	
\put(179,135){\mydot}	
\put(180,135){\mydot}	
\put(181,136){\mydot}	
\put(182,136){\mydot}	
\put(183,135){\mydot}	
\put(184,135){\mydot}	
\put(185,135){\mydot}	
\put(186,134){\mydot}	
\put(187,133){\mydot}	
\put(188,132){\mydot}	
\put(189,133){\mydot}	
\put(190,131){\mydot}	
\put(191,132){\mydot}	
\put(192,132){\mydot}	
\put(193,135){\mydot}	
\put(194,138){\mydot}	
\put(195,139){\mydot}	
\put(196,141){\mydot}	
\put(197,142){\mydot}	
\put(198,144){\mydot}	
\put(199,145){\mydot}	
\put(200,146){\mydot}	
\put(201,144){\mydot}	
\put(202,144){\mydot}	
\put(203,144){\mydot}	
\put(204,143){\mydot}	
\put(205,144){\mydot}	
\put(206,144){\mydot}	
\put(207,144){\mydot}	
\put(208,143){\mydot}	
\put(209,141){\mydot}	
\put(210,139){\mydot}	
\put(211,138){\mydot}	
\put(212,136){\mydot}	
\put(213,135){\mydot}	
\put(214,132){\mydot}	
\put(215,129){\mydot}	
\put(216,129){\mydot}	
\put(217,128){\mydot}	
\put(218,128){\mydot}	
\put(219,128){\mydot}	
\put(220,129){\mydot}	
\put(221,126){\mydot}	
\put(222,125){\mydot}	
\put(223,124){\mydot}	
\put(224,124){\mydot}	
\put(225,122){\mydot}	
\put(226,119){\mydot}	
\put(227,132){\mydot}	
\put(228,123){\mydot}	
\put(229,123){\mydot}	
\put(230,123){\mydot}	
\put(231,124){\mydot}	
\put(232,124){\mydot}	
\put(233,122){\mydot}	
\put(234,121){\mydot}	
\put(235,122){\mydot}	
\put(236,122){\mydot}	
\put(237,122){\mydot}	
\put(238,122){\mydot}	
\put(239,122){\mydot}	
\put(240,121){\mydot}	
\put(241,120){\mydot}	
\put(242,120){\mydot}	
\put(243,120){\mydot}	
\put(244,121){\mydot}	
\put(245,121){\mydot}	
\put(246,122){\mydot}	
\put(247,123){\mydot}	
\put(248,124){\mydot}	
\put(249,127){\mydot}	
\put(250,129){\mydot}	
\put(251,131){\mydot}	
\put(252,131){\mydot}	
\put(254,130){\mydot}	
\put(255,131){\mydot}	
\put(256,131){\mydot}	
\put(257,129){\mydot}	
\put(258,128){\mydot}	
\put(259,127){\mydot}	
\put(260,127){\mydot}	
\put(261,126){\mydot}	
\put(262,125){\mydot}	
\put(263,123){\mydot}	
\put(264,123){\mydot}	
\put(265,120){\mydot}	
\put(266,118){\mydot}	
\put(267,117){\mydot}	
\put(268,117){\mydot}	
\put(269,115){\mydot}	
\put(270,114){\mydot}	
\put(271,114){\mydot}	
\put(272,111){\mydot}	
\put(273,112){\mydot}	
\put(274,112){\mydot}	
\put(275,112){\mydot}	
\put(276,112){\mydot}	
\put(277,111){\mydot}	
\put(278,112){\mydot}	
\put(279,111){\mydot}	
\put(280,110){\mydot}	
\put(281,114){\mydot}	
\put(282,118){\mydot}	
\put(283,116){\mydot}	
\put(284,117){\mydot}	
\put(285,116){\mydot}	
\put(286,115){\mydot}	
\put(287,114){\mydot}	
\put(288,114){\mydot}	
\put(289,114){\mydot}	
\put(290,114){\mydot}	
\put(291,114){\mydot}	
\put(292,113){\mydot}	
\put(293,113){\mydot}	
\put(294,112){\mydot}	
\put(295,111){\mydot}	
\put(296,111){\mydot}	
\put(297,111){\mydot}	
\put(298,112){\mydot}	
\put(299,112){\mydot}	
\put(300,112){\mydot}	
\put(301,113){\mydot}	
\put(302,114){\mydot}	
\put(303,116){\mydot}	
\put(304,118){\mydot}	
\put(305,120){\mydot}	
\put(306,121){\mydot}	
\put(307,123){\mydot}	
\put(308,124){\mydot}	
\put(309,122){\mydot}	
\put(310,122){\mydot}	
\put(311,121){\mydot}	
\put(312,120){\mydot}	
\put(313,120){\mydot}	
\put(314,119){\mydot}	
\put(315,119){\mydot}	
\put(316,118){\mydot}	
\put(317,117){\mydot}	
\put(318,116){\mydot}	
\put(319,115){\mydot}	
\put(320,114){\mydot}	
\put(321,114){\mydot}	
\put(322,113){\mydot}	
\put(323,110){\mydot}	
\put(324,108){\mydot}	
\put(325,104){\mydot}	
\put(326,108){\mydot}	
\put(327,108){\mydot}	
\put(328,107){\mydot}	
\put(329,106){\mydot}	
\put(330,105){\mydot}	
\put(331,105){\mydot}	
\put(332,104){\mydot}	
\put(333,101){\mydot}	
\put(334,102){\mydot}	
\put(335,104){\mydot}	
\put(336,113){\mydot}	
\put(337,108){\mydot}	
\put(338,111){\mydot}	
\put(339,108){\mydot}	
\put(340,107){\mydot}	
\put(341,107){\mydot}	
\put(342,107){\mydot}	
\put(343,106){\mydot}	
\put(344,106){\mydot}	
\put(345,106){\mydot}	
\put(346,107){\mydot}	
\put(347,106){\mydot}	
\put(348,106){\mydot}	
\put(349,106){\mydot}	
\put(350,106){\mydot}	
\put(351,105){\mydot}	
\put(352,105){\mydot}	
\put(353,105){\mydot}	
\put(354,105){\mydot}	
\put(355,107){\mydot}	
\put(356,108){\mydot}	
\put(357,109){\mydot}	
\put(358,110){\mydot}	
\put(359,111){\mydot}	
\put(360,112){\mydot}	
\put(361,113){\mydot}	
\put(362,113){\mydot}	
\put(363,115){\mydot}	
\put(364,115){\mydot}	
\put(365,113){\mydot}	
\put(366,113){\mydot}	
\put(367,112){\mydot}	
\put(368,112){\mydot}	
\put(369,112){\mydot}	
\put(370,111){\mydot}	
\put(371,109){\mydot}	
\put(372,108){\mydot}	
\put(373,107){\mydot}	
\put(374,105){\mydot}	
\put(375,104){\mydot}	
\put(376,103){\mydot}	
\put(377,103){\mydot}	
\put(378,102){\mydot}	
\put(379,100){\mydot}	
\put(380,100){\mydot}	
\put(381,101){\mydot}	
\put(382,100){\mydot}	
\put(383,99){\mydot}	
\put(384,99){\mydot}	
\put(385,98){\mydot}	
\put(386,97){\mydot}	
\put(387,98){\mydot}	
\put(388,107){\mydot}	
\put(389,107){\mydot}	
\put(390,108){\mydot}	
\put(391,107){\mydot}	
\put(392,104){\mydot}	
\put(393,104){\mydot}	
\put(394,105){\mydot}	
\put(395,104){\mydot}	
\put(396,104){\mydot}	
\put(397,103){\mydot}	
\put(398,103){\mydot}	
\put(399,103){\mydot}	
\put(400,102){\mydot}	
\put(401,102){\mydot}	
\put(402,101){\mydot}	
\put(403,99){\mydot}	
\put(404,98){\mydot}	
\put(405,101){\mydot}	
\put(406,103){\mydot}	
\put(407,102){\mydot}	
\put(408,104){\mydot}	
\put(409,107){\mydot}	
\put(410,108){\mydot}	
\put(411,111){\mydot}	
\put(412,116){\mydot}	
\put(413,116){\mydot}	
\put(414,116){\mydot}	
\put(415,117){\mydot}	
\put(416,116){\mydot}	
\put(417,114){\mydot}	
\put(418,114){\mydot}	
\put(419,113){\mydot}	
\put(420,112){\mydot}	
\put(421,111){\mydot}	
\put(422,110){\mydot}	
\put(423,109){\mydot}	
\put(424,108){\mydot}	
\put(425,106){\mydot}	
\put(426,104){\mydot}	
\put(427,104){\mydot}	
\put(428,101){\mydot}	
\put(429,101){\mydot}	
\put(430,101){\mydot}	
\put(431,101){\mydot}	
\put(432,101){\mydot}	
\put(433,100){\mydot}	
\put(434,99){\mydot}	
\put(435,97){\mydot}	
\put(436,95){\mydot}	
\put(437,96){\mydot}	
\put(438,99){\mydot}	
\put(439,104){\mydot}	
\put(440,107){\mydot}	
\put(441,102){\mydot}	
\put(442,100){\mydot}	
\put(443,100){\mydot}	
\put(444,100){\mydot}	
\put(445,99){\mydot}	
\put(446,99){\mydot}	
\put(447,99){\mydot}	
\put(448,98){\mydot}	
\put(449,99){\mydot}	
\put(450,98){\mydot}	
\put(451,98){\mydot}	
\put(452,97){\mydot}	
\put(453,96){\mydot}	
\put(454,95){\mydot}	
\put(455,99){\mydot}	
\put(456,101){\mydot}	
\put(457,104){\mydot}	
\put(458,105){\mydot}	
\put(459,106){\mydot}	
\put(460,107){\mydot}	
\put(461,108){\mydot}	
\put(462,109){\mydot}	
\put(463,116){\mydot}	
\put(464,114){\mydot}	
\put(465,113){\mydot}	
\put(466,112){\mydot}	
\put(467,112){\mydot}	
\put(468,111){\mydot}	
\put(469,110){\mydot}	
\put(470,109){\mydot}	
\put(471,108){\mydot}	
\put(472,107){\mydot}	
\put(473,106){\mydot}	
\put(474,106){\mydot}	
\put(475,105){\mydot}	
\put(476,103){\mydot}	
\put(477,101){\mydot}	
\put(478,99){\mydot}	
\put(479,98){\mydot}	
\put(480,95){\mydot}	
\put(481,97){\mydot}	
\put(482,96){\mydot}	
\put(483,96){\mydot}	
\put(484,95){\mydot}	
\put(485,95){\mydot}	
\put(486,94){\mydot}	
\put(487,93){\mydot}	
\put(488,92){\mydot}	
\put(489,91){\mydot}	
\put(490,94){\mydot}	
\put(491,98){\mydot}	
\put(492,98){\mydot}	
\put(493,96){\mydot}	
\put(494,96){\mydot}	
\put(495,95){\mydot}	
\put(496,95){\mydot}	
\put(497,95){\mydot}	
\put(498,94){\mydot}	
\put(499,94){\mydot}	
\put(500,95){\mydot}	
\put(501,94){\mydot}	
\put(502,94){\mydot}	
\put(503,94){\mydot}	
\put(504,93){\mydot}	
\put(505,92){\mydot}	
\put(506,90){\mydot}	
\put(507,90){\mydot}	
\put(508,93){\mydot}	
\put(509,93){\mydot}	
\put(510,94){\mydot}	
\put(511,95){\mydot}	
\put(512,96){\mydot}	
\put(513,96){\mydot}	
\put(514,97){\mydot}	
\put(515,97){\mydot}	
\put(516,100){\mydot}	
\put(517,99){\mydot}	
\put(518,98){\mydot}	
\put(519,99){\mydot}	
\put(520,99){\mydot}	
\put(521,99){\mydot}	
\put(522,100){\mydot}	
\put(523,100){\mydot}	
\put(524,100){\mydot}	
\put(525,100){\mydot}	
\put(526,100){\mydot}	
\put(527,99){\mydot}	
\end{picture}
\end{center}

\caption{F$_0$ Trace for Bamileke Dschang Utterance:
`child and child and ... '}
\end{figure*}
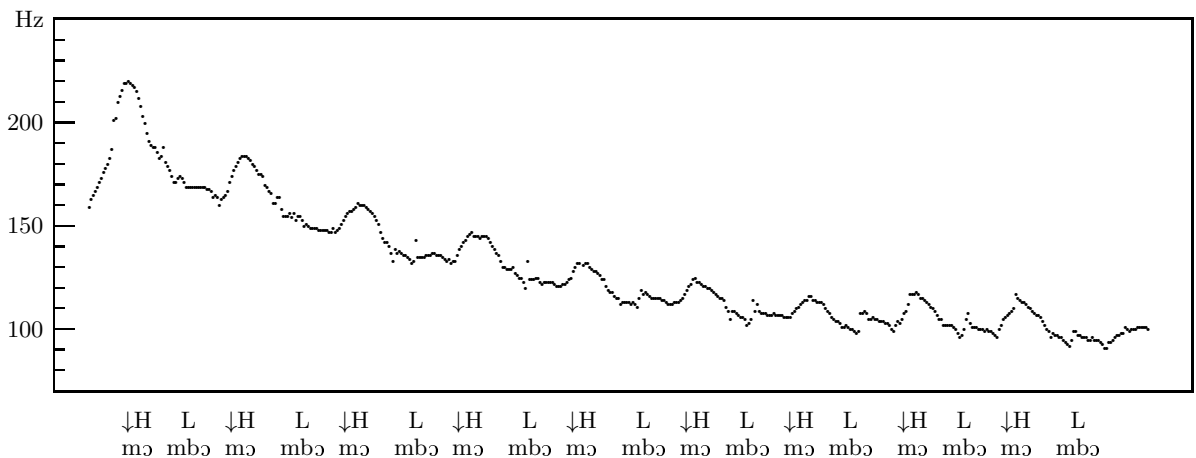
The question addressed in this paper concerns how we should relate
pitch contours to tone sequences.

This paper is divided into four main sections,
summarised in turn below.
\begin{description}
\item[Tone Transcription]
In this section I present the problem of relating
sequences of F$_0$ values to tone transcriptions.
I argue that Hidden Markov Models are unsuited to
the task and I demonstrate the importance of having
a computational tool which allows phonologists to
experiment with F$_0$ scaling parameters.
\item[{\boldmath F$_0$ Scaling}]
This section gives a mathematical basis for a
general approach to F$_0$ scaling which, it is hoped,
will be applicable to any tone language.
I derive an F$_0$ prediction function from first
principles and show how the model of \shortciteA{Liberman93}
for the Nigerian language Igbo is a special case.
\item[{\boldmath Tone and F$_0$ in Bamileke Dschang}]
\mbox{Here I} present some data from my own fieldwork and give
a statistical analysis, using the same technique used
by Liberman et al.  I then show how the general model
of the previous section is instantiated for this
language.  This demonstrates the versatility of the
general model, since it can be applied to two very
different tone languages.
\item[Implementations]
This section provides two non-deterministic techniques
for transcribing an F$_0$ string.  The first method
uses a genetic algorithm while the second method
uses simulated annealing.  The performance of both
implementations is evaluated and compared on a range
of artificial and real data.  Finally, I give some examples
of multiple, automatically-generated transcriptions of the same
F$_0$ data.
\end{description}

\section{TONE TRANSCRIPTION}
\subsection{Generation and Recognition}

A promising way of generating contours from tone sequences is to
specify one or more pitch targets per tone and then to interpolate
between the targets; the task then becomes one of providing a suitable
sequence of targets \cite{PierrehumbertBeckman88}.
It is perhaps less clear
how we should go about recognising tone sequences from pitch contours.
Hidden Markov Models (HMMs) \shortcite{Huang90} offer a powerful
statistical approach to this problem, though it is unclear how they
could be used to recognise the units of interest to phonologists.
HMMs do not encode timing information in a way that would allow them
to output, say, one tone per syllable (or vowel).
Moreover, the same section
of a pitch contour may correspond to either H or L tones.  For
example, a H between two Hs looks just like an L between two Ls.
There is no principled upper bound on the amount of context that needs
to be inspected in order to resolve the ambiguity, leading to a
multiplication of state information required by the HMM and problems
for training it.

In the present context, the emphasis is not on automatic speech
recognition but on a tool to support phonologists working with tone.
As we shall see in the next section, once the phonologist has
identified the salient location to measure the `F$_0$ value' of a
syllable (or some other phonological unit), the task will be
to automatically map a string of these values to a string of tones.

\subsection{A Tool for Phonologists}

Connell and Ladd have devised a set of heuristics for identifying key
points in an F$_0$ contour to record F$_0$ values
\cite[21ff]{ConnellLadd90}.
In the absence of a
program which enshrines these heuristics, it was decided to develop
a system for producing a tone transcription from a sequence of
F$_0$ values.
Apart from the obvious benefits of automating the process, such as speed
and accuracy, it could show up cases where there is more than one
possible tone transcription, possibly with different parameter settings for
the F$_0$ scaling function.  Having the {\it set}
of tone transcriptions that are
compatible with an utterance has considerable
value to an analyst searching for invariances in the tonal assignments
to individual morphemes.

To exemplify this point, it is worth considering a recent example
where an alternative transcription of some data proved valuable in
providing a fresh analysis of the data.  In their analyses of tone
in Bamileke Dschang, Hyman gives the
transcription in (\ref{ex:machete-dogs}a) while Stewart gives
the one in (\ref{ex:machete-dogs}b), for the phrase meaning
\ling{machete of dogs}.
\begin{ex}
\label{ex:machete-dogs}
\begin{subexamples}
\item
\trans{\`{\nj}\negsp\nj\v{\i} m\`\schwa m\downstep bh\'\baru}
--- \cite[50]{Hyman85}
\item
\trans{\`{\nj}\negsp\nj\`{\i}\upstep\'{\ }\downstep m\`\schwa mbh\'\baru}
--- \cite[200]{Stewart93}
\end{subexamples}
\end{ex}
These two possibilities exist because of different
F$_0$ \emph{scaling parameters}.  These parameters determine
the way in which the different tones are scaled relative to each other
and to the speaker's pitch range.  This is illustrated in
(\ref{ex:hyman-interp}), adapting Hyman's earlier notation
\cite{Hyman79}.

\begin{ex}
\label{ex:hyman-interp}
\begin{subexamples}
\item
Hyman:
\trans{\`{\nj}\negsp\nj\v{\i} m\`\schwa m\downstep bh\'\baru} \\
\begin{tabular}[t]{|llllll|}
\hline
\trans{\`{\nj}} & \trans{\nj\`{\i}} & \trans{\'{\i}} &
\trans{m\`\schwa} & \trans{\downstep} & \trans{mbh\'\baru} \\
L & L & H & L & \downstep & H \\ \hline
3 & 3 & 1  & 3 &           & 1 \\
0 & 0 & 0  & 0 & 1         & 1 \\ \hline
3 & 3 & 1  & 3 &           & 2 \\ \hline
\end{tabular}
\vspace{3mm}

\item
Stewart:
\trans{\`{\nj}\negsp\nj\`{\i}\upstep\'{\ }\downstep m\`\schwa mbh\'\baru} \\
\begin{tabular}[t]{|lllllll|}
\hline
\trans{\`{\nj}} & \trans{\nj\`{\i}} & \trans{\upstep} &
\trans{\'{\i}} & \trans{\downstep} & \trans{m\`\schwa} &
\trans{mbh\'\baru} \\
L & L & \upstep & H & \downstep & L & H \\ \hline
2 & 2 &         & 1  &           & 2 & 1 \\
1 & 1 & 0       & 0  & 1         & 1 & 1 \\ \hline
3 & 3 &         & 1  &           & 3 & 2 \\ \hline
\end{tabular}
\end{subexamples}
\end{ex}

\hspace*{-\parindent}
Example (\ref{ex:hyman-interp}) displays a kind of phonetic
interpretation function.
Immediately below the two rows of tones we see a row of numbers
corresponding to the tones.  For Hyman, L=3 and H=1, while for
Stewart, L=2 and H=1.  Observe in Hyman's example that a rising
tone---symbolised by a wedge above the \ling{i}---is modelled as an LH
sequence in keeping with standard practice in African tone analysis.

The second row of numbers corresponds to downstep (\downstep) and
upstep (\upstep).  For Hyman's model, this row begins at 0 and is
increased by 1 for each downstep encountered.  For Stewart's model,
this row begins at 1 and is increased by 1 for each
downstep encountered and decreased by 1 for each upstep encountered.
The two rows are summed vertically to give the last row of numbers.
Observe that the last rows of Stewart's and Hyman's models are
identical.

The parameter which distinguishes the
two approaches is \emph{partial} vs.~\emph{total} downstep.
Hyman treats Dschang as a \emph{partial downstep language},
i.e.~where \downstep H appears as a mid tone (with respect to
the material to its left).
Stewart treats it as a \emph{total downstep language},
i.e.~where \downstep H appears as an L tone (with respect to
the material to its left).

While Hyman and Stewart present rather different analyses of
rather different looking transcriptions, we can see that they
are really analyzing the same data, given the above interpretation
function.  Therefore, phonologists who do not wish to limit themselves
to the transcriptions which result from certain parameter settings in
the phonetic interpretation function would be better off working
directly with number sequences like the last row in
(\ref{ex:hyman-interp}).  This paper describes a tool which
lets them do just that.

\section{F$_0$ SCALING}
\label{scaling}

Consider again the F$_0$ contour in Figure 1.  In particular, note
that the F$_0$ decay seems to be to a non-zero asymptote, and that H
and L appear to have different asymptotes which we symbolise as $h$
and $l$ respectively.  These observations are
clearer in Figure 2, which (roughly speaking) displays the peaks and
valleys from Figure 1.

\begin{figure}[ht]
\begin{center}
\setlength{\unitlength}{0.3mm}
\begin{picture}(230,170)(30,80)
\put(40,80){\framebox(200,160){}}
\multiput(40,108)(5,0){40}{\line(1,0){2.5}}
\multiput(40,94)(5,0){40}{\line(1,0){2.5}}
\put(245,108){\makebox(0,0)[l]{\small $h$}}
\put(245,94){\makebox(0,0)[l]{\small $l$}}
\put(35,100){\makebox(0,0)[r]{100}}
\put(35,150){\makebox(0,0)[r]{150}}
\put(35,200){\makebox(0,0)[r]{200}}
\put(35,240){\makebox(0,0)[r]{Hz}}
\put(185,210){\scriptsize LEGEND}
\put(185,200){\scriptsize $\frown$\ = H}
\put(185,190){\scriptsize $\smile$\ = L}
\put(180,185){\framebox(50,35){}}
\put(40,90){\line(1,0){2}}
\put(40,100){\line(1,0){5}}
\put(40,110){\line(1,0){2}}
\put(40,120){\line(1,0){2}}
\put(40,130){\line(1,0){2}}
\put(40,140){\line(1,0){2}}
\put(40,150){\line(1,0){5}}
\put(40,160){\line(1,0){2}}
\put(40,170){\line(1,0){2}}
\put(40,180){\line(1,0){2}}
\put(40,190){\line(1,0){2}}
\put(40,200){\line(1,0){5}}
\put(40,210){\line(1,0){2}}
\put(40,220){\line(1,0){2}}
\put(40,230){\line(1,0){2}}
%
\put( 50,219){\makebox(0,0)[t]{\scriptsize$\frown$}}
\put( 60,168){\makebox(0,0)[b]{\scriptsize$\smile$}}
\put( 70,183){\makebox(0,0)[t]{\scriptsize$\frown$}}
\put( 80,150){\makebox(0,0)[b]{\scriptsize$\smile$}}
\put( 90,160){\makebox(0,0)[t]{\scriptsize$\frown$}}
\put(100,136){\makebox(0,0)[b]{\scriptsize$\smile$}}
\put(110,144){\makebox(0,0)[t]{\scriptsize$\frown$}}
\put(120,123){\makebox(0,0)[b]{\scriptsize$\smile$}}
\put(130,131){\makebox(0,0)[t]{\scriptsize$\frown$}}
\put(140,115){\makebox(0,0)[b]{\scriptsize$\smile$}}
\put(150,122){\makebox(0,0)[t]{\scriptsize$\frown$}}
\put(160,107){\makebox(0,0)[b]{\scriptsize$\smile$}}
\put(170,113){\makebox(0,0)[t]{\scriptsize$\frown$}}
\put(180,105){\makebox(0,0)[b]{\scriptsize$\smile$}}
\put(190,118){\makebox(0,0)[t]{\scriptsize$\frown$}}
\put(200,100){\makebox(0,0)[b]{\scriptsize$\smile$}}
\put(210,113){\makebox(0,0)[t]{\scriptsize$\frown$}}
\put(220, 95){\makebox(0,0)[b]{\scriptsize$\smile$}}
\end{picture}

\caption{Asymptotic Behaviour of F$_0$}
\end{center}
\end{figure}
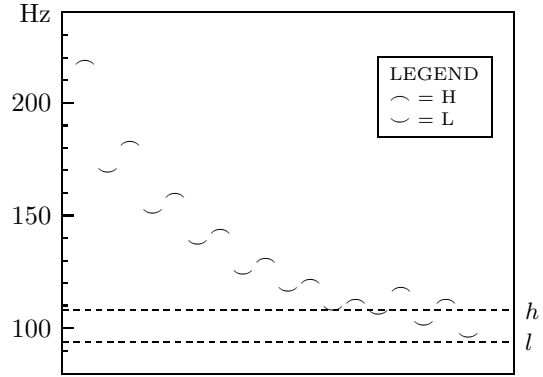

Although this is admittedly a rather artificial example,
it remains true that there is no principled upper limit
on the number of downsteps that can occur in an utterance
\cite[540]{Clements79}, and so the asymptotic behaviour of
F$_0$ scaling still needs to be addressed.

Now suppose that we have a sequence $T$ of tones where
$t_i$ is the $i$th tone (H or L) and a sequence $X$ of
F$_0$ values where $x_i$ is the F$_0$ value corresponding
to $t_i$.
Then we would like a formula which predicts $x_i$ given
$x_{i-1}$, $t_{i}$ and $t_{i-1}$ ($i>1$).  We express this as
follows:

\[
x_i = {\cal P}_{t_{i-1} t_i}(x_{i-1})
\]
The question, now, is what should this function look like?
Suppose for sake of argument that the ratio of L to the immediately
preceding H in Figure 2 is constant, with respect to the
baselines for H and L, namely $h$ and $l$.  Then we have:

\begin{eqnarray*}
\frac{x_i - l}{x_{i-1} - h} &=& c
\end{eqnarray*}
More generally, suppose that we have a sequence of two arbitrary
tones.  Ignoring the possibility of downstep for the present,
we have a static two-tone system where HH and LL sequences
are level and sequences like HLHLHL are realised as simple oscillation
between two pitches.  We can write the following formula, where
$\tbar{i}=h$ if $t_i = H$ and
$\tbar{i}=l$ if $t_i = L$.
\begin{eqnarray*}
\frac{x_i - \tbar{i}}{x_{i-1} - \tbar{i-1}} &=&
\frac{\tbar{i}}{\tbar{i-1}} \\
\Rightarrow
x_i &=& \frac{\tbar{i}}{\tbar{i-1}}.x_{i-1}
\end{eqnarray*}
The situation becomes more interesting when we allow for
downdrift and downstep.
\emph{Downdrift} is the automatic lowering of the second of
two H tones when an L intervenes, so HLH is realised
as \mbox{[\Hdash \Ldash \Mdash]}
rather than as \mbox{[\Hdash \Ldash \Hdash]},
while \emph{downstep} is the lowering of the second of two
tones when an intervening L is lost,
so H\downstep H is realised as \mbox{[\Hdash \Mdash]}
\cite{HymanSchuh74}.
Bamileke Dschang has downstep but not downdrift while
Igbo has downdrift but only very limited downstep.
Now we define $\tbar{i}=h$ if $t_i = $H, \downstep H and
$\tbar{i}=l$ if $t_i = $L, \downstep L.
Generalising our equation once more, we have the following,
where $R$ is a factor called the \emph{transition ratio}.
\begin{eqnarray*}
\frac{x_i - \tbar{i}}{x_{i-1} - \tbar{i-1}} &=&
\frac{\tbar{i}}{\tbar{i-1}}\Rm{t_{i-1} t_i} \\
\Rightarrow
x_i = \Pm{t_{i-1} t_i}(x_{i-1})
 &=& \frac{\tbar{i}}{\tbar{i-1}}\Rm{t_{i-1} t_i}.x_{i-1} \\
&+& \tbar{i}(1 - \Rm{t_{i-1} t_i})
\end{eqnarray*}
Now I shall show how this general equation relates to the
equations for Igbo \shortcite[151]{Liberman93}, reproduced below:

\begin{examples}
\item\label{ex:liberman}
\begin{tabular}[t]{ll}
HH & $x_i = x_{i-1}$ \\
HL & $x_i = (Fl/h)x_{i-1} + l(1-F)$ \\
LH & $x_i = (h/l)x_{i-1}$ \\
LL & $x_i = Fx_{i-1} + l(1-F)$ \\
H\downstep H & $x_i = Dx_{i-1} + h(1-D)$
\end{tabular}
\end{examples}
$\cal P$ can be instantiated to the
set of equations in (\ref{ex:liberman}) by setting $R$
as follows:

\begin{tabular}{|r|ccc|}
\multicolumn{1}{l}{} &
\multicolumn{3}{c}{$t_i$} \\ \cline{2-4}
\multicolumn{1}{c|}{$t_{i-1}$} & H & L & \downstep H \\ \hline
H, \downstep H & 1 & F & D \\
L & 1 & F & -- \\ \hline
\end{tabular}
\hspace{1cm}
\(\begin{array}{l}
0 < F < 1 \\
0 < D < 1
\end{array}\)
\vspace{3mm}

It will be helpful to introduce one more level of
generality.  $\cal P$ relates {\it adjacent}
F$_0$ values, but we would also like to relate {\it non-adjacent}
values, given the sequence of intervening tones.
Suppose that $T=t_0\cdots t_n$ is a tone sequence where
the F$_0$ value of $t_0$ is $x$.  Then we shall write the
F$_0$ value of $t_n$ as $\Pm{T}(x)$.  By repeated applications
of $\cal P$ we can write down the following expression for $\Pm{T}$:
\[
\Pm{T}(x) =
\frac{\tbar{n}}{\tbar{0}}\Rm{T}.x
+ \tbar{n}(1-\Rm{T})
\]
where $\Rm{T} = \prod_{k=1}^n \Rm{t_{k-1}t_k}$, $n>2$.
Now, suppose that $S=s_0\cdots s_m$ and
$T=t_0\cdots t_n$ are tone sequences and
that \mbox{$\sbar{0} = \tbar{0}$},
\mbox{$\sbar{m} = \tbar{n}$} and \mbox{$\Pm{S} = \Pm{T}$}.
Then it is straightforward to show that $R_S = R_T$.
Notice also that if $\Pm{T}(x) = x$ for all $x$
and if $\tbar{0} = \tbar{n}$ then $\Rm{T} = 1$.
These results will be useful in the next section.

Finally, it is worth comparing $\cal P$ with
Hyman's and Stewart's interpretation functions which were illustrated
in (\ref{ex:hyman-interp}).  As pointed out already,
Hyman's is a partial downstep
model while Stewart's is a total downstep model.
Partial and total downstep
can be visualised as follows, where the dotted lines indicate the
abstract {\it register} inside which tones are scaled, and where
downstep corresponds to lowering of the register.

\hspace*{-\parindent}
\begin{minipage}[t]{0pt}
\vspace*{30mm}
\includegraphics{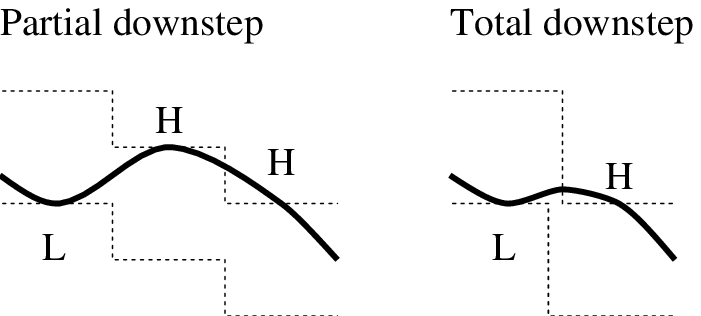}
\vspace*{5mm}
\end{minipage}

\hspace*{-\parindent}
Observe that for partial downstep, it is necessary to have two
downsteps before a high tone is at the level of a preceding low,
while for total downstep, it is only necessary to have a single
downstep for a high tone to be at the same level as the preceding
low.  We can express these observations about partial and total
downstep in the model as follows.  For partial downstep,
we have $\P{L\downstep H\downstep H}(x) = x$ while
for total downstep we have $\P{L\downstep H}(x) = x$.
For both of these equations
we are forced to have $h=l$ which does not seem to be
empirically justifiable in view of the data in Figure 1.
It might be argued that this indicates a flaw in the model
being presented here,
since partial and total downstep are widely attested in
the literature on tone languages.  Unfortunately, it is
not possible in general to provide a model for
partial or total downstep which
permits distinct asymptotes for H and L.\footnote{
To see why this is so for the case of
total downstep, suppose that such a model did exist,
and so $l < h$.  Let $x \in [l,h)$, a valid F$_0$ value for a low tone.
Now, whatever interpretation function $\cal P'$ we use, we still
require that $\Pp{L\downstep H}(x) = x$ by definition
of total downstep, which means that
there is now a high tone with a F$_0$ value less than $h$.
But $h$ is the asymptote below which no high tones should
ever be realised, and so we have a contradiction.  The case
for partial downstep follows similarly.}
Therefore, to the extent that Figure 1 is typical of tone
languages in having different H and L asymptotes,
one must conclude that total and partial downstep
are qualitative terms only.  However, they may yet re-emerge
in the model under a different guise, as we shall see later.

The effect of the distinction between partial and
total downstep is to allow different transcriptions of the
same string, as we saw in (\ref{ex:hyman-interp}).
In general, we have the following mapping between
transcriptions under the two views of downstep:

\begin{examples}
\item\label{ex:mapping}
\begin{tabular}[t]{ccc|ccc}
partial && total & partial && total \\
HH & --- & HH & L\ds H & --- & LH \\
HL & --- & H\ds L & L\ds L & --- & L\ds L\\
LH & --- & L\us H & H\us H & --- & H\us H\\
LL & --- & LL & H\us L & --- & HL \\
H\ds H & --- & H\ds H & L\us L & --- & L\us L \\
\end{tabular}
\end{examples}

\hspace*{-\parindent}
It is clear that changing from one view of downstep to the other
amounts to adding and deleting \ds\ and \us\ while leaving the
tones themselves unchanged.  Thus, the model
admits both transcription schemes that result from the
two views of downstep, and another besides, as shown
later in (\ref{equiv}).

This concludes the discussion of the F$_0$ prediction
function.  In the next section I shall investigate the
phonetic interpretation of tone in Bamileke Dschang, and
determine the values of $R$ for this language.

\section{{\boldmath TONE AND F$_0$ IN \\ BAMILEKE DSCHANG}}

In a recent field trip to Western Cameroon to study the
Bamileke Dschang\footnote{
Bamileke Dschang is a grassfields Bantu language spoken in the
Western Province of Cameroon.  The name `Bamileke'
(pron: \transx{[b\scripta\stress mileke]}) represents both an
ethnic grouping and a language cluster; Dschang
(pron: \transx{[t\esh\scripta\eng]}) is an important town
around which one of the Bamileke languages is spoken.  The data here
is from the Bafou dialect.}
noun associative construction, I was able
to collect a small amount of data relating to F$_0$ scaling
throughout a particular informant's pitch range.  Following
Liberman et al., voice pitch was
varied by getting the informant to speak at different volumes
and by adjusting the recording level appropriately.  However,
rather than asking the informant to imagine speaking to a subject
at different distances, I controlled the volume by having the
informant wear headphones and played white noise from a detuned
radio.  Thus, I could set the informant's voice pitch by using
the volume control on my radio.  My hypothesis is that this
technique produces more consistent volume (and hence, pitch scaling)
over long utterances and may make informants less self-conscious
about speaking loudly than simply asking them
to imagine speaking to subjects at various distances away.
Measurements were taken from the following data.

\begin{ex}
\label{ex:data}
\begin{description}
\item[HH]
\trans{\'{\scripta} \yogh \baru \'{\openo}  s\'{\schwa}\eng\
t\'e n\yogh \baru \'{\openo}  t\'{\schwa}\eng\
t\'e n\yogh \baru \'{\openo}  t\'u\eng\
t\'e n\yogh \baru \'{\openo}  k\'{\scripta}p
t\'e n\yogh \baru \'{\openo}  k\'{\openo}p} \\
\ling{He saw the bird before he saw the hat before
he saw the basket before he saw the pipe before he saw the cup}
\item[LL]
\trans{\`{\scripta}p\`{\scripta}k} --- \ling{side, half}
\item[L\downstep H, HL] \hfill\\
\trans{\`e\downstep s\'{\openo} mb\`{\openo} \`e\downstep s\'{\openo}
mb\`{\openo}  ... \`e\downstep s\'{\openo}} \\
\ling{jealousy and jealousy and ... jealousy} \\
\trans{l\`{\schwa}\downstep p\'{\schwa} mb\`{\openo}
l\`{\schwa}\downstep p\'{\schwa} mb\`{\openo}
... l\`{\schwa}\downstep p\'{\schwa}} \\
\ling{breast and breast and ... breast} \\
\trans{m\`e\downstep v\'{\niepsilon}t mb\`{\openo}
m\`e\downstep v\'{\niepsilon}t mb\`{\openo}  ...
m\`e\downstep v\'{\niepsilon}t} \\
\ling{oil and oil and ... oil}\\
\trans{\downstep m\'{\openo} mb\`{\openo}
\downstep m\'{\openo}  mb\`{\openo}  ... \downstep m\'{\openo}} \\
\ling{child and child and ... child}
\end{description}
\end{ex}

Regrettably, the LL data was only available from isolated disyllables,
and other sequences such as LH and H\downstep H were not available
at all.  However, from the F$_0$ data for the above utterances
we can hypothesise the behaviour of these unseen sequences, and this
can be tested in subsequent empirical work.
The results for utterances involving HH and LL sequences
are displayed in Figure 3, while results for
L\downstep H and HL are displayed in Figure 4.

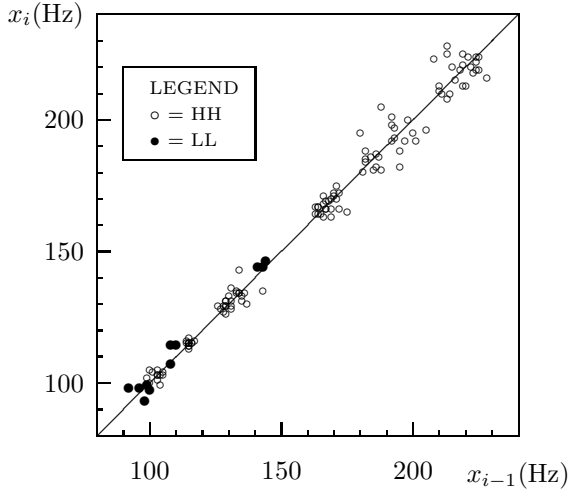
\begin{figure}[ht]
\begin{center}
\setlength{\unitlength}{0.35mm}
\begin{picture}(250,170)(50,70)
\put(80,80){\framebox(160,160){}}
\put(80,80){\line(1,1){160}}
\put(100,70){\makebox(0,0)[t]{100}}
\put(150,70){\makebox(0,0)[t]{150}}
\put(200,70){\makebox(0,0)[t]{200}}
\put(75,100){\makebox(0,0)[r]{100}}
\put(75,150){\makebox(0,0)[r]{150}}
\put(75,200){\makebox(0,0)[r]{200}}
\put(75,240){\makebox(0,0)[r]{$x_i$(Hz)}}
\put(240,70){\makebox(0,0)[t]{$x_{i-1}$(Hz)}}
\put(100,210){\scriptsize LEGEND}
\put(100,200){\scriptsize \hhlab\ = HH}
\put(100,190){\scriptsize \lllab\ = LL}
\put(90,185){\framebox(50,35){}}
\put(80,90){\line(1,0){2}}
\put(80,100){\line(1,0){5}}
\put(80,110){\line(1,0){2}}
\put(80,120){\line(1,0){2}}
\put(80,130){\line(1,0){2}}
\put(80,140){\line(1,0){2}}
\put(80,150){\line(1,0){5}}
\put(80,160){\line(1,0){2}}
\put(80,170){\line(1,0){2}}
\put(80,180){\line(1,0){2}}
\put(80,190){\line(1,0){2}}
\put(80,200){\line(1,0){5}}
\put(80,210){\line(1,0){2}}
\put(80,220){\line(1,0){2}}
\put(80,230){\line(1,0){2}}
\put(90,80){\line(0,1){2}}
\put(100,80){\line(0,1){5}}
\put(110,80){\line(0,1){2}}
\put(120,80){\line(0,1){2}}
\put(130,80){\line(0,1){2}}
\put(140,80){\line(0,1){2}}
\put(150,80){\line(0,1){5}}
\put(160,80){\line(0,1){2}}
\put(170,80){\line(0,1){2}}
\put(180,80){\line(0,1){2}}
\put(190,80){\line(0,1){2}}
\put(200,80){\line(0,1){5}}
\put(210,80){\line(0,1){2}}
\put(220,80){\line(0,1){2}}
\put(230,80){\line(0,1){2}}
%
\put(98,93){\ll}
\put(99,99){\ll}
\put(100,97){\ll}
\put(92,98){\ll}
\put(96,98){\ll}
\put(108,107){\ll}
\put(108,114){\ll}
\put(110,114){\ll}
\put(141,144){\ll}
\put(144,146){\ll}
\put(143,144){\ll}
%
\put(115,115){\hh}
\put(115,114){\hh}
\put(114,116){\hh}
\put(116,115){\hh}
\put(115,115){\hh}
\put(115,117){\hh}
\put(117,116){\hh}
\put(116,115){\hh}
\put(115,115){\hh}
\put(115,114){\hh}
\put(114,115){\hh}
\put(115,113){\hh}

\put(129,131){\hh}
\put(131,131){\hh}
\put(131,129){\hh}
\put(129,131){\hh}
\put(131,128){\hh}
\put(128,127){\hh}
\put(127,128){\hh}
\put(128,129){\hh}
\put(129,129){\hh}
\put(129,126){\hh}
\put(126,129){\hh}
\put(129,129){\hh}

\put(165,164){\hh}
\put(164,164){\hh}
\put(164,167){\hh}
\put(167,169){\hh}
\put(169,163){\hh}
\put(163,164){\hh}
\put(164,167){\hh}
\put(167,166){\hh}
\put(166,163){\hh}
\put(163,167){\hh}
\put(167,166){\hh}

\put(193,193){\hh}
\put(193,197){\hh}
\put(197,192){\hh}
\put(192,192){\hh}
\put(192,201){\hh}
\put(201,192){\hh}
\put(192,198){\hh}
\put(198,200){\hh}
\put(200,195){\hh}
\put(195,188){\hh}
\put(188,205){\hh}
\put(205,196){\hh}

\put(214,210){\hh}
\put(210,211){\hh}
\put(211,210){\hh}
\put(210,213){\hh}
\put(213,228){\hh}
\put(228,216){\hh}
\put(216,215){\hh}
\put(215,220){\hh}
\put(220,213){\hh}
\put(213,208){\hh}
\put(208,223){\hh}
\put(223,218){\hh}

\put(100,100){\hh}
\put(100,105){\hh}
\put(105,104){\hh}
\put(104,103){\hh}
\put(103,103){\hh}
\put(103,103){\hh}
\put(103,105){\hh}
\put(105,103){\hh}
\put(103,101){\hh}
\put(101,104){\hh}
\put(104,99){\hh}
\put(99,102){\hh}

\put(137,130){\hh}
\put(130,133){\hh}
\put(133,134){\hh}
\put(134,143){\hh}
\put(143,135){\hh}
\put(135,133){\hh}
\put(133,135){\hh}
\put(135,131){\hh}
\put(131,136){\hh}
\put(136,134){\hh}
\put(134,134){\hh}
\put(134,134){\hh}

\put(169,170){\hh}
\put(170,171){\hh}
\put(171,170){\hh}
\put(170,172){\hh}
\put(172,172){\hh}
\put(172,166){\hh}
\put(166,168){\hh}
\put(168,169){\hh}
\put(169,166){\hh}
\put(166,171){\hh}
\put(171,175){\hh}
\put(175,165){\hh}

\put(182,185){\hh}
\put(185,181){\hh}
\put(181,180){\hh}
\put(180,195){\hh}
\put(195,182){\hh}
\put(182,184){\hh}
\put(184,186){\hh}
\put(186,187){\hh}
\put(187,186){\hh}
\put(186,182){\hh}
\put(182,188){\hh}
\put(188,181){\hh}

\put(218,219){\hh}
\put(219,221){\hh}
\put(221,224){\hh}
\put(224,219){\hh}
\put(219,213){\hh}
\put(213,225){\hh}
\put(225,219){\hh}
\put(219,225){\hh}
\put(225,224){\hh}
\put(224,224){\hh}
\put(224,222){\hh}
\put(222,220){\hh}
\end{picture}
\end{center}

\caption{Plot of $x_{i-1}$ vs $x_i$ for HH, LL}
\end{figure}

\begin{figure}[ht]
\begin{center}
\setlength{\unitlength}{0.35mm}
\begin{picture}(250,170)(50,70)
\put(80,80){\framebox(160,160){}}
\put(80,80){\line(1,1){160}}
\put(100,70){\makebox(0,0)[t]{100}}
\put(150,70){\makebox(0,0)[t]{150}}
\put(200,70){\makebox(0,0)[t]{200}}
\put(75,100){\makebox(0,0)[r]{100}}
\put(75,150){\makebox(0,0)[r]{150}}
\put(75,200){\makebox(0,0)[r]{200}}
\put(75,240){\makebox(0,0)[r]{$x_i$(Hz)}}
\put(240,70){\makebox(0,0)[t]{$x_{i-1}$(Hz)}}
\put(100,210){\scriptsize LEGEND}
\put(100,200){\scriptsize \ldhlab\ = L!H}
\put(100,190){\scriptsize \hllab\ = HL}
\put(90,185){\framebox(50,35){}}
\put(80,90){\line(1,0){2}}
\put(80,100){\line(1,0){5}}
\put(80,110){\line(1,0){2}}
\put(80,120){\line(1,0){2}}
\put(80,130){\line(1,0){2}}
\put(80,140){\line(1,0){2}}
\put(80,150){\line(1,0){5}}
\put(80,160){\line(1,0){2}}
\put(80,170){\line(1,0){2}}
\put(80,180){\line(1,0){2}}
\put(80,190){\line(1,0){2}}
\put(80,200){\line(1,0){5}}
\put(80,210){\line(1,0){2}}
\put(80,220){\line(1,0){2}}
\put(80,230){\line(1,0){2}}
\put(90,80){\line(0,1){2}}
\put(100,80){\line(0,1){5}}
\put(110,80){\line(0,1){2}}
\put(120,80){\line(0,1){2}}
\put(130,80){\line(0,1){2}}
\put(140,80){\line(0,1){2}}
\put(150,80){\line(0,1){5}}
\put(160,80){\line(0,1){2}}
\put(170,80){\line(0,1){2}}
\put(180,80){\line(0,1){2}}
\put(190,80){\line(0,1){2}}
\put(200,80){\line(0,1){5}}
\put(210,80){\line(0,1){2}}
\put(220,80){\line(0,1){2}}
\put(230,80){\line(0,1){2}}
%
\put(175,137){\hl}
\put(150,122){\hl}
\put(133,107){\hl}
\put(119,102){\hl}
\put(116,97){\hl}
\put(108,94){\hl}
\put(108,95){\hl} %
\put(109,93){\hl} %
\put(224,167){\hl}
\put(200,156){\hl}
\put(175,136){\hl}
\put(156,127){\hl}
\put(140,118){\hl}
\put(129,109){\hl}
\put(119,103){\hl}
\put(120,102){\hl} %
\put(111,95){\hl}  %
\put(211,160){\hl}
\put(178,143){\hl}
\put(159,129){\hl}
\put(140,118){\hl}
\put(129,113){\hl}
\put(125,108){\hl}
\put(121,107){\hl} %
\put(116,103){\hl} %
\put(113,94){\hl}  %
\put(135,112){\hl}
\put(122,104){\hl}
\put(114,103){\hl}
\put(110,97){\hl}
\put(106,91){\hl}  %
\put(219,168){\hl}
\put(183,150){\hl}
\put(160,136){\hl}
\put(144,123){\hl}
\put(131,115){\hl}
\put(122,107){\hl}
\put(113,105){\hl} %
\put(118,100){\hl} %
\put(113,95){\hl}  %
%
\put(137,150){\ldh}
\put(122,133){\ldh}
\put(107,119){\ldh}
\put(102,116){\ldh}
\put(97,108){\ldh}
\put(94,108){\ldh} %
\put(95,109){\ldh} %
\put(93,112){\ldh} %
\put(205,224){\ldh}
\put(167,200){\ldh}
\put(156,175){\ldh}
\put(136,156){\ldh}
\put(127,140){\ldh}
\put(118,129){\ldh}
\put(109,119){\ldh}
\put(103,120){\ldh} %
\put(102,111){\ldh} %
\put(190,211){\ldh}
\put(160,178){\ldh}
\put(143,159){\ldh}
\put(129,140){\ldh}
\put(118,129){\ldh}
\put(113,125){\ldh}
\put(108,121){\ldh}
\put(107,116){\ldh} %
\put(103,113){\ldh} %
\put(112,122){\ldh}
\put(104,114){\ldh}
\put(103,110){\ldh}
\put(97,106){\ldh}
\put(168,183){\ldh}
\put(150,160){\ldh}
\put(136,144){\ldh}
\put(123,131){\ldh}
\put(115,122){\ldh}
\put(107,113){\ldh}
\put(105,118){\ldh} %
\put(100,113){\ldh} %
\end{picture}
\end{center}

\caption{Plot of $x_{i-1}$ vs $x_i$ for L\downstep H, HL}
\end{figure}

The regression equations obtained from these data are displayed
in (\ref{ex:regression}), where the number of occurrences of each
tone sequence is given in parentheses after the sequence.  The
third column gives the standard error for the gradient and intercept.
\begin{ex}
\label{ex:regression}
{\small
\begin{tabular}[t]{|lll|}
\hline
Tone      & Regression & Standard \\
Sequence  & Equation   & Error \\ \hline
HH (119) & $x_i = 0.99 x_{i-1}  +  0.91$ & 0.012, 5.0 \\
LL (11) & $x_i = 1.02 x_{i-1}  -  1.39$ & 0.057, 3.6 \\
HL (40) & $x_i = 0.65 x_{i-1}  +  25.0$ & 0.015, 3.1 \\
L\downstep H (38) & $x_i = 1.10 x_{i-1}  + 0.54$ & 0.026, 4.3 \\ \hline
\end{tabular}
}
\end{ex}
{}From this, we conclude that HL is the only sequence with an
intercept significantly different from zero,
and that $x_i = x_{i-1}$ for HH and LL sequences.
We also conclude that
$\R{HH} = \R{LL} = \R{L\downstep H} = 1$, ($l/h = 1.1$)
and $\R{HL} = 0.72$.
This last value will be referred to as the quantity $d$.
We also see that $l=88$Hz and $h=96$Hz.
Fortunately, these figures are sufficient to
determine the $R$ values for all other pairs of tones
in Bamileke Dschang.

A further observation is that Bamileke Dschang does not have
downdrift, and so there
is no F$_0$ difference across HLH and LHL sequences.
This is evident in Figure 5.
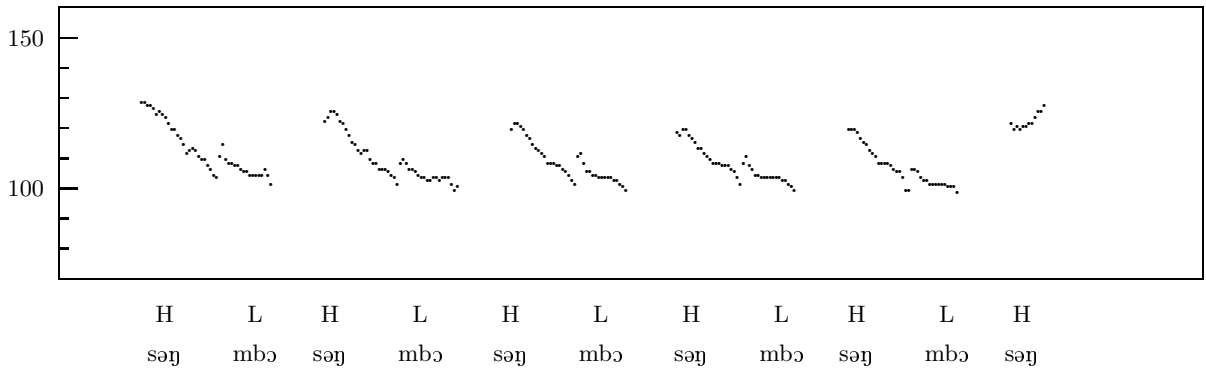
\begin{figure*}
\setlength{\unitlength}{0.4mm}
\begin{picture}(400,150)(-30,40)
\put(0,70){\framebox(380,90){}}
\put(0,80){\line(1,0){3}}
\put(0,90){\line(1,0){3}}
\put(0,100){\line(1,0){6}} \put(-5,100){\makebox(0,0)[r]{\small 100}}
\put(0,110){\line(1,0){3}}
\put(0,120){\line(1,0){3}}
\put(0,130){\line(1,0){3}}
\put(0,140){\line(1,0){3}}
\put(0,150){\line(1,0){6}} \put(-5,150){\makebox(0,0)[r]{\small 150}}

\pair{35}{\tH}{\SENG}
\pair{65}{\tL}{\MBO}
\pair{90}{\tH}{\SENG}
\pair{120}{\tL}{\MBO}
\pair{150}{\tH}{\SENG}
\pair{180}{\tL}{\MBO}
\pair{210}{\tH}{\SENG}
\pair{240}{\tL}{\MBO}
\pair{265}{\tH}{\SENG}
\pair{295}{\tL}{\MBO}
\pair{320}{\tH}{\SENG}

\put(26,128){\mydot}	
\put(27,128){\mydot}	
\put(28,127){\mydot}	
\put(29,127){\mydot}	
\put(30,126){\mydot}	
\put(31,124){\mydot}	
\put(32,125){\mydot}	
\put(33,124){\mydot}	
\put(34,123){\mydot}	
\put(35,121){\mydot}	
\put(36,119){\mydot}	
\put(37,119){\mydot}	
\put(38,117){\mydot}	
\put(39,116){\mydot}	
\put(40,114){\mydot}	
\put(41,111){\mydot}	
\put(42,112){\mydot}	
\put(43,113){\mydot}	
\put(44,112){\mydot}	
\put(45,110){\mydot}	
\put(46,109){\mydot}	
\put(47,109){\mydot}	
\put(48,107){\mydot}	
\put(49,106){\mydot}	
\put(50,104){\mydot}	
\put(51,103){\mydot}	
\put(52,110){\mydot}	
\put(53,114){\mydot}	
\put(54,109){\mydot}	
\put(55,108){\mydot}	
\put(56,108){\mydot}	
\put(57,107){\mydot}	
\put(58,107){\mydot}	
\put(59,106){\mydot}	
\put(60,105){\mydot}	
\put(61,105){\mydot}	
\put(62,104){\mydot}	
\put(63,104){\mydot}	
\put(64,104){\mydot}	
\put(65,104){\mydot}	
\put(66,104){\mydot}	
\put(67,106){\mydot}	
\put(68,104){\mydot}	
\put(69,101){\mydot}	
\put(87,122){\mydot}	
\put(88,123){\mydot}	
\put(89,125){\mydot}	
\put(90,125){\mydot}	
\put(91,124){\mydot}	
\put(92,122){\mydot}	
\put(93,121){\mydot}	
\put(94,119){\mydot}	
\put(95,117){\mydot}	
\put(96,115){\mydot}	
\put(97,114){\mydot}	
\put(98,112){\mydot}	
\put(99,111){\mydot}	
\put(100,112){\mydot}	
\put(101,112){\mydot}	
\put(102,109){\mydot}	
\put(103,108){\mydot}	
\put(104,108){\mydot}	
\put(105,106){\mydot}	
\put(106,106){\mydot}	
\put(107,106){\mydot}	
\put(108,105){\mydot}	
\put(109,104){\mydot}	
\put(110,103){\mydot}	
\put(111,101){\mydot}	
\put(112,108){\mydot}	
\put(113,109){\mydot}	
\put(114,108){\mydot}	
\put(115,106){\mydot}	
\put(116,106){\mydot}	
\put(117,105){\mydot}	
\put(118,104){\mydot}	
\put(119,103){\mydot}	
\put(120,103){\mydot}	
\put(121,102){\mydot}	
\put(122,102){\mydot}	
\put(123,103){\mydot}	
\put(124,103){\mydot}	
\put(125,102){\mydot}	
\put(126,103){\mydot}	
\put(127,103){\mydot}	
\put(128,103){\mydot}	
\put(129,101){\mydot}	
\put(130,99){\mydot}	
\put(131,100){\mydot}	
\put(149,119){\mydot}	
\put(150,121){\mydot}	
\put(151,121){\mydot}	
\put(152,120){\mydot}	
\put(153,119){\mydot}	
\put(154,117){\mydot}	
\put(155,116){\mydot}	
\put(156,114){\mydot}	
\put(157,113){\mydot}	
\put(158,112){\mydot}	
\put(159,111){\mydot}	
\put(160,110){\mydot}	
\put(161,108){\mydot}	
\put(162,108){\mydot}	
\put(163,108){\mydot}	
\put(164,107){\mydot}	
\put(165,107){\mydot}	
\put(166,106){\mydot}	
\put(167,105){\mydot}	
\put(168,104){\mydot}	
\put(169,102){\mydot}	
\put(170,101){\mydot}	
\put(171,110){\mydot}	
\put(172,111){\mydot}	
\put(173,108){\mydot}	
\put(174,105){\mydot}	
\put(175,105){\mydot}	
\put(176,104){\mydot}	
\put(177,104){\mydot}	
\put(178,103){\mydot}	
\put(179,103){\mydot}	
\put(180,103){\mydot}	
\put(181,103){\mydot}	
\put(182,103){\mydot}	
\put(183,102){\mydot}	
\put(184,102){\mydot}	
\put(185,101){\mydot}	
\put(186,100){\mydot}	
\put(187,99){\mydot}	
\put(204,118){\mydot}	
\put(205,117){\mydot}	
\put(206,119){\mydot}	
\put(207,119){\mydot}	
\put(208,117){\mydot}	
\put(209,116){\mydot}	
\put(210,115){\mydot}	
\put(211,113){\mydot}	
\put(212,113){\mydot}	
\put(213,111){\mydot}	
\put(214,110){\mydot}	
\put(215,109){\mydot}	
\put(216,108){\mydot}	
\put(217,108){\mydot}	
\put(218,108){\mydot}	
\put(219,107){\mydot}	
\put(220,107){\mydot}	
\put(221,107){\mydot}	
\put(222,106){\mydot}	
\put(223,105){\mydot}	
\put(224,103){\mydot}	
\put(225,101){\mydot}	
\put(226,108){\mydot}	
\put(227,110){\mydot}	
\put(228,107){\mydot}	
\put(229,106){\mydot}	
\put(230,104){\mydot}	
\put(231,104){\mydot}	
\put(232,103){\mydot}	
\put(233,103){\mydot}	
\put(234,103){\mydot}	
\put(235,103){\mydot}	
\put(236,103){\mydot}	
\put(237,103){\mydot}	
\put(238,103){\mydot}	
\put(239,102){\mydot}	
\put(240,102){\mydot}	
\put(241,101){\mydot}	
\put(242,100){\mydot}	
\put(243,99){\mydot}	
\put(261,119){\mydot}	
\put(262,119){\mydot}	
\put(263,119){\mydot}	
\put(264,118){\mydot}	
\put(265,116){\mydot}	
\put(266,115){\mydot}	
\put(267,114){\mydot}	
\put(268,112){\mydot}	
\put(269,111){\mydot}	
\put(270,110){\mydot}	
\put(271,108){\mydot}	
\put(272,108){\mydot}	
\put(273,108){\mydot}	
\put(274,108){\mydot}	
\put(275,107){\mydot}	
\put(276,106){\mydot}	
\put(277,105){\mydot}	
\put(278,105){\mydot}	
\put(279,103){\mydot}	
\put(280,99){\mydot}	
\put(281,99){\mydot}	
\put(282,106){\mydot}	
\put(283,106){\mydot}	
\put(284,105){\mydot}	
\put(285,103){\mydot}	
\put(286,102){\mydot}	
\put(287,102){\mydot}	
\put(288,101){\mydot}	
\put(289,101){\mydot}	
\put(290,101){\mydot}	
\put(291,101){\mydot}	
\put(292,101){\mydot}	
\put(293,101){\mydot}	
\put(294,100){\mydot}	
\put(295,100){\mydot}	
\put(296,100){\mydot}	
\put(297,98){\mydot}	
\put(315,121){\mydot}	
\put(316,119){\mydot}	
\put(317,120){\mydot}	
\put(318,119){\mydot}	
\put(319,120){\mydot}	
\put(320,120){\mydot}	
\put(321,121){\mydot}	
\put(322,121){\mydot}	
\put(323,123){\mydot}	
\put(324,125){\mydot}	
\put(325,125){\mydot}	
\put(326,127){\mydot}	
\end{picture}

\caption{F$_0$ Trace for `bird and bird and ... '}
\end{figure*}
Therefore, we can write $\P{HLH}(x)=x$, and by a result we showed above,
$\R{HL}.\R{LH}=1$.  Given that $\R{HL} = d$
it follows that $\R{LH} = \frac{1}{d}$.

Concerning downstep, I shall assume that the magnitude
of downstep is independent of the tones on either side,
and so $\P{HL\downstep H} = \P{H\downstep H}
= \P{L\downstep L} = \P{LH\downstep L}$.  A separate
instrumental study supports this hypothesis
\cite{BirdStegen93a}.  Therefore, we have
$\R{st} = \frac{1}{d}\R{s\ds t} = d\R{s\us t}$,
where $s$ is any tone and $t$ is H or L.

Finally, it is important to briefly consider upstep,
since it has been used in some analyses of Bamileke
Dschang (e.g.~Stewart's).  Given that upstep and downstep
are intended as inverses of each other, we have the
identities
$\Pm{s\downarrow t\uparrow t}
= \Pm{st}
= \Pm{s\uparrow t\downarrow t}$,
with $s, t$ as before.
We now have a complete table for $R$:

\hspace*{-\parindent}
\begin{tabular}{|r|cc|cc|cc|}
\multicolumn{1}{l}{} &
\multicolumn{6}{c}{$t_i$} \\ \cline{2-7}
\multicolumn{1}{c|}{$t_{i-1}$} &
H & L &
\downstep H & \downstep L &
\upstep H & \upstep L \\ \hline

H, \downstep H, \upstep H &
$1$ & $d$ & $d$ & $d^2$ & $d^{-1}$ & 1 \\

L, \downstep L, \upstep L &
$d^{-1}$ & $1$ & $1$ & $d$ & $d^{-2}$ & $d^{-1}$ \\ \hline
\end{tabular}
\vspace{3mm}

\hspace*{-\parindent}
Observe the symmetries in this table.  The configuration
of four $R$ values that we find when $t_i$ is not
downstepped or upstepped (the first two columns) is
reproduced in the columns for downstep (multiplied by $d$)
and in the columns for upstep (divided by $d$).

Note also that the above table is dependent upon
how the data in (\ref{ex:data}) was transcribed.
Suppose that we had not used
repetitions of HL\downstep H (a~transcription scheme
based on partial downstep) but H\downstep LH (a scheme
based on total downstep).
Then we would have had $\R{H\downstep L} = d$ and $\R{LH} = 1$.
Accordingly, the table for $R$ would be as follows:

\hspace*{-\parindent}
\begin{tabular}{|r|cc|cc|cc|}
\multicolumn{1}{l}{} &
\multicolumn{6}{c}{$t_i$} \\ \cline{2-7}
\multicolumn{1}{c|}{$t_{i-1}$} &
H & L &
\downstep H & \downstep L &
\upstep H & \upstep L \\ \hline

H, \downstep H, \upstep H &
$1$ & $1$ & $d$ & $d$ & $d^{-1}$ & $d^{-1}$ \\

L, \downstep L, \upstep L &
$1$ & $1$ & $d$ & $d$ & $d^{-1}$ & $d^{-1}$ \\ \hline
\end{tabular}
\vspace{3mm}

The fact that we have two possible tables for $R$ is
no cause for alarm.  Recall that the transition between
two tones $t_{i-1}$ and $t_i$ also involves the factor
$\tbar{i}/\tbar{i-1}$.  This factor is manifested in
tone transitions according to the following pattern:

\hspace*{-\parindent}
\begin{tabular}{|r|cc|cc|cc|}
\multicolumn{1}{l}{} &
\multicolumn{6}{c}{$t_i$} \\ \cline{2-7}
\multicolumn{1}{c|}{$t_{i-1}$} &
H & L &
\downstep H & \downstep L &
\upstep H & \upstep L \\ \hline

H, \downstep H, \upstep H &
$1$ & $l/h$ & $1$ & $l/h$ & $1$ & $l/h$ \\

L, \downstep L, \upstep L &
$h/l$ & $1$ & $h/l$ & $1$ & $h/l$ & $1$ \\ \hline
\end{tabular}
\vspace{3mm}

\hspace*{-\parindent}
I therefore conclude that the presence of more than one
table for $R$ indicates an interplay between $R$
values and the ratio $h/l$.  This raises an
interesting question.  Suppose we have two tone sequences
$T = t_0 \cdots t_n$ and $T' = t'_0 \cdots t'_n$,
and two interpretation functions $\cal P$ and $\cal P'$
based on $R$ and $R'$ respectively.
Then under what circumstances is the phonetic interpretation
of both sequences the same under their respective
interpretation functions?  A sufficient condition for them
to be the same is that $\tbar{i} = \tpbar{i}$
and that $R_{t_{i-1}t_i} = R'_{t'_{i-1}t'_i}$.
The reader can check that these conditions are met by
the mapping in (\ref{ex:mapping}) and the two tables for $R$ given above.
Note that this observation holds for the model in general, not just
for the specialised version of the model as applied
to Bamileke Dschang.

It can also be shown that $R$ is completely
determined once $\R{HL}$ is specified.  A possible
characterisation of total vs.~partial downstep now arises:
if $\R{HL}=1$ then we have total downstep,
but if $\R{HL}=d<1$ then we have partial downstep.
However, the interpretation of these terms must necessarily
be different from the standard interpretation, since I have
shown that the standard interpretation is not compatible with
the present model.

This concludes the discussion of F$_0$ scaling in Bamileke Dschang.
I shall now present the implementations.

\section{IMPLEMENTATIONS}

In this section, I show how it is possible to get two programs to
produce a sequence of tones $T$ (i.e.~a tone transcription)
given a sequence of $n$ F$_0$ values $X$.
The programs make crucial use of the prediction function $\cal P$
in evaluating candidate tone transcriptions.

Both programs involve search, and in general,
the aim in searching is to discover the
values for $x_1, \ldots, x_n$ so
as to optimise the value of a specified
evaluation function $f(x_1, \ldots, x_n)$.
When $f$ has many local optima, deterministic methods such
as hill-climbing perform poorly.  This is because they
terminate in a local optimum and the particular one found
depends heavily on the starting point in the search, and there
is usually no way of choosing a good starting point.

Exhaustive search for the global optimum is not an option
when the search space is prohibitively large.  In the
present context, say for
a sequence of 20 tones, the search space contains $6^{20} \approx
10^{15}$ possible tone transcriptions, and for each of these there are
thousands of possible parameter settings, too large a search space for
exhaustive search in a reasonable amount of computation time.

{\em Non-deterministic search methods} have been devised as a way of
tackling large-scale combinatorial optimisation problems,
problems that involve finding optima of functions of discrete variables.
These methods are only designed to yield an approximate solution, but
they do so in a reasonable amount of computation time.  The best
known such methods are
genetic search \cite{Goldberg89}
and annealing search \cite{Laarhoven87}.
Recently, annealing search has been successfully applied to the learning
of phonological constraints expressed as finite-state automata
\cite{Ellison92b}.  In the following sections I describe
a genetic algorithm and an annealing algorithm for the
tone transcription problem.

\subsection{A Genetic Algorithm}

For a cogent introduction to genetic search and an explanation of why
it works, the reader is referred to \shortcite{South93}.
Before presenting the version of the algorithm used in
the implementation, I shall informally define
the key data types it uses along with the standard operations on
those types.

\begin{description}
\item[gene]
  A linear encoding of a solution.
  In the present setting, it is an array of $n$ tones, where each tone
  is one of H, \downstep H, \upstep H, L, \downstep L or \upstep L.
  A gene also contains 16 bit encodings of the parameters $h$, $l$
  and $d$.  These encodings were scaled to be floating point numbers in
  the range $[90,110]$ for $h$, $[70,100]$ for $l$ and $[0.6,0.9]$ for $d$.
\item[gene pool]
  An array of genes, $P$.  One of the search parameters
  is the size of $P$, known as the \emph{population}.
  The gene pool is renewed each generation, and the number of
  generations is another search parameter.
\item[evaluation]
  A measure of the fitness of a gene as a solution to the problem.
  Suppose that $X$ is the sequence of F$_0$ values we wish to
  transcribe.  Suppose also that $T$ is a particular gene.
  The the evaluation function is as follows:
  \[
  {\cal E}_X(T) = \frac{1}{n}
  \sum_{i=2}^{n} ({\cal P}_{t_{i-1} t_i}(x_{i-1}) - x_i)^2
  \]
\item[crossover]
  This is an operation which takes two genes and produces
  a single gene as the result.
  Suppose that $A=a_1\cdots a_n$ and $B=b_1\cdots b_n$.  Then
  the crossover function $C_r$ is defined as follows, where
  $r$ is the (randomly selected) \emph{crossover point}
  ($0 \leq r \leq n$).
  \begin{eqnarray*}
  && C_r(a_1\cdots a_r a_{r+1}\cdots a_n,
  b_1\cdots b_r b_{r+1}\cdots b_n) \\ &=&
  a_1\cdots a_r b_{r+1}\cdots b_n
  \end{eqnarray*}
  In other words, the genes $A$ and $B$ are cut at a position
  determined by $r$ and the first part of $A$ is spliced with the
  second part of $B$ to create a new gene.
  Crossover builds in the idea that good genes tend to produce good
  offspring.  To see why this is so, suppose that the transcription
  contained in the first part of $A$ is relatively good while the rest
  is poor, while the transcription contained in the first part of $B$
  is poor and the rest is relatively good.  Then the offspring
  containing the first part of $A$ and the second part of $B$ will be
  an improvement on both $A$ and $B$; other possible offspring
  from $A$ and $B$ will
  be significantly worse and may not survive to the next generation.
  The program performs this kind of crossover for
  the parameters $h$, $l$ and $d$, employing independent crossover
  points for each, and randomising the argument order in $C_r$ so
  that the high order bits in the offspring are equally likely to
  come from either parent.

  An extension to crossover allows more than one crossing point.
  The current model permits an {\em arbitrary number of crossing points}
  for crossover on the transcription string.  The resulting gene
  is optimal since we choose the crossing points in such a way as to
  minimise $({\cal P}_{t_{i-1} t_i}(x_{i-1}) - x_i)^2$ at each position.
  In developing the system, exploiting the decomposability of the
  evaluation function in this way caused a significant improvement
  in system performance over the version which used simple crossover.
\item[breeding]
  For each generation, we create a new gene pool from the
  previous one.  Each new gene is created by mating
  the best of three randomly chosen
  genes with the best of three other randomly chosen genes.
\item[mutation]
  In order to maintain some \emph{genetic diversity} and an
  element of randomness throughout the search (rather than just
  in the initial configuration), a further
  operation is applied to each gene in every generation.
  With a certain probability (known as the \emph{mutation
  probability}), for each gene $T$ and each tone in $T$, the tone is
  randomly set to any of the six possible tones.  Likewise,
  the parameter encodings are mutated.
  The mutation rate is set to 0.005 but raised to 0.5 for
  a single generation if the
  evaluation of the best gene is no improvement on the evaluation
  of the best gene ten generations earlier.
  The best gene is never mutated.
\end{description}

The building blocks of genetic search discussed above are
structured into the following algorithm, expressed in
pseudo-Pascal:

\begin{tabbing}
xx\=xx\=xx\=xx\=xx\=xx\kill
{\bf procedure genetic\_search}                        \\
begin                                                  \\
\> {\bf initialise} Pool, NewPool;                     \\
\> for g := 1 to generations do                        \\
\> begin                                               \\
\>\> if {\bf good\_performance}(10) then               \\
\>\>\> mutation\_rate := 0.005;                        \\
\>\> else                                              \\
\>\>\> mutation\_rate := 0.5;                          \\
\>\> NewPool[1] := {\bf find\_best\_gene}(Pool);       \\
\>\> for n := 2 to population do                       \\
\>\> begin                                             \\
\>\>\> gene1 := {\bf best\_of\_three}(Pool);           \\
\>\>\> gene2 := {\bf best\_of\_three}(Pool);           \\
\>\>\> NewPool[n] := {\bf crossover}(gene1, gene2);    \\
\>\>\> {\bf mutate}(NewPool[n], mutation\_rate);       \\
\>\> end                                               \\
\>\> Pool := NewPool;                                  \\
\>\> {\bf evaluate}(Pool);                             \\
\> end                                                 \\
\> write {\bf find\_best\_gene}(Pool);                 \\
end
\end{tabbing}

\hspace*{-\parindent}
The main loop is executed for each generation.  Each time
through this loop, the program
checks performance over the last ten generations
and if performance has been good, the mutation rate stays low,
otherwise it is changed to high.
Then it copies the best gene to the new pool.
Now we reach the inner loop,
which selects two genes, performs crossover, and mutates the
result.  Next, the current pool is updated, an evaluation is
performed, and the program continues with the next generation.
Once all the generations have been completed, the program
displays the best gene from the final population and terminates.

\subsection{An Annealing Algorithm}

As with genetic algorithms, simulated annealing \cite{Laarhoven87}
is a combinatorial optimisation
technique based on an analogy with a natural process.
{\em Annealing} is the heating and slow cooling of a solid
which allows the formation of regular crystalline structure
having a minimum of excess energy.
In its early stages when the temperature is high,
annealing search resembles random search.
There is so much free energy in the system that
a transition to a higher energy state is highly probable.
As the temperature decreases the search begins to resemble
hill-climbing.  Now there is much less free energy and
so transitions to higher energy states are less and
less likely.
In what follows, I explain some of
the parameters of annealing search as used in the
current implementation.
\begin{description}
\item[temperature]
At the start of the search the temperature, $t$ is
set to 1.  During the search, the temperature is
reduced at a rate set by the `cooling rate' parameter,
until it reaches a value less than $10^{-6}$.
\item[perturbation]
At each step of the search, the current state is
perturbed by an amount which depends on the temperature.
The temperature determines the fraction of the search
space that is covered by a single perturbation step.  For a tone sequence
of length $n$,
we randomly reset the worst $n.t$ tones according to
$({\cal P}_{t_{i-1} t_i}(x_{i-1}) - x_i)^2$.  For the
parameters we proceed as follows, here exemplified for $h$.
First, set
$\rho = t(h_{\mbox{\scriptsize max}} - h_{\mbox{\scriptsize min}})$.
Now, add to $h$ a random number in the range $[-\rho, \rho]$
and check that the result is still in the range
$[h_{\mbox{\scriptsize min}}, h_{\mbox{\scriptsize max}}]$.

\item[equilibrium]
At each temperature, the system is required to
reach `thermal equilibrium' before the temperature
is lowered.  In the present context,
equilibrium is reached if no more than one of the
last eight perturbations yielded a new state that
was accepted.

\item[free energy function]
This is the amount of available energy for
transitions to higher energy states.
In the current system, it is the distribution
$-1000.t.log(p)$, where $p$ is
a uniform random variable in the range $(0,1]$.
If the energy difference $\Delta$ between an old and a new
state is less than the available energy, then
the transition is accepted.  The factor of 1000 is
intended to scale the energy distribution to typical
values of the evaluation function.
\end{description}
Now the algorithm itself is presented:

\begin{tabbing}
xx\=xx\=xx\=xx\=xx\= xx\kill
{\bf procedure annealing\_search}                      \\
begin                                                  \\
\> {\bf initialise} Trans, NewTrans, BestTrans;        \\
\> {\bf randomise} Trans;                              \\
\> t := 1;                                             \\
\> while t $>$ 0.000001 do                             \\
\> begin                                               \\
\>\> repeat                                            \\
\>\>\> NewTrans := {\bf perturb}(Trans, t);            \\
\>\>\> $\Delta$ := {\bf evaluate}(NewTrans)            \\
\>\>\>\> $-$ {\bf evaluate}(Trans);                    \\
\>\>\> if $\Delta < 0$ or                               \\
\>\>\>\> exp($-\Delta$/1000.t) $>$ random(0,1) then    \\
\>\>\>\> Trans := NewTrans;                              \\
\>\>\> if {\bf evaluate}(Trans) $<$ {\bf evaluate}(BestTrans) \\
\>\>\>\> BestTrans := Trans;                             \\
\>\> until {\bf equilibrium\_reached};                   \\
\>\> Trans := BestTrans;                                 \\
\>\> temperature := temperature / 1.2;                   \\
\> end                                                   \\
\> write Trans;                                          \\
end
\end{tabbing}

\hspace*{-\parindent}
The program is made up of two loops.  The outer loop simply
iterates through the temperature range, beginning with a
temperature of 1 and steadily decreasing it until it gets
very close to zero.  The nested loop performs the task of
reaching thermal equilibrium at each temperature.
The first step is to perturb the previous transcription to
make a new one.  Notice that the temperature $t$ is a parameter
of the perturb function.  Next, the difference $\Delta$ between
the old and new evaluations is calculated.  If the new transcription
has a better evaluation than the old one, then $\Delta$ is negative.
Next, the program accepts the new transcription if (i) $\Delta$ is negative
or (ii) $\Delta$ is positive and there is sufficient free energy in
the system to allow the worse transcription to be accepted.
Finally, we check if the new transcription is better than the best
transcription found so far (BestTrans) and if so, we set BestTrans
to be the new transcription.  Once equilibrium is reached, the current
transcription is set to be the best transcription found so far, and
the search continues.

\subsection{Performance Results}

Both the genetic and annealing search algorithms have been implemented
in \cpp.  In this section, the performance of the two implementations
is compared.  Performance statistics are based on 1,200 executions
of each program.  Search parameters were set so that each execution took
around 5 seconds on a Sun Sparc 10.
Three performance trials were undertaken.

\hspace*{-\parindent}
{\bf Trial 1: Artificial Data.}
In the first trial, both programs generated random sequences of
tones, then computed the corresponding F$_0$ sequence using
$\cal P$, then set about transcribing the F$_0$ sequence.
Since these sequences were ideal, the best possible evaluation
for a transcription was zero.  The performance of the programs
could then be measured to see how close they came to
finding the optimal solution.  Each program was tested on
F$_0$ sequences of length 5, 10, 15 and 20.  For each length,
each program transcribed 100 randomly-generated sequences.
The results are displayed in Figure 6.  Each pair of bars
corresponds to a given transcription length.  The left member
of each pair is for the genetic search program, while the
right member is for the annealing search program.

\begin{figure}
\vspace{65mm}
\includegraphics{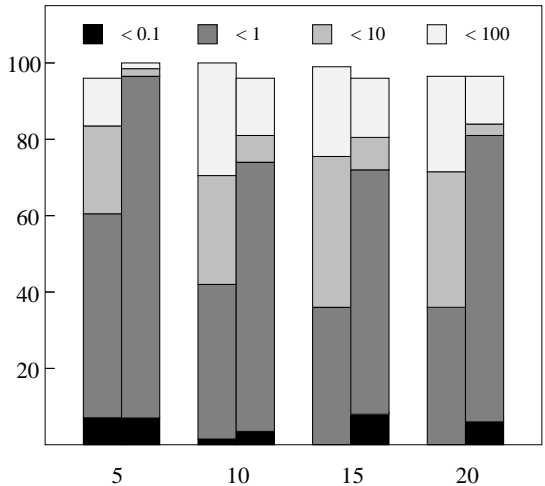}
\caption{Performance results (no upstep)}
\end{figure}

The heavily shaded bars corresponding to evaluations less than 1 are the
most important.  These indicate the number of times out of 100 that
the programs found a transcription with an evaluation less than 1.
This evaluation means that the average
of the squared difference between the predicted
F$_0$ values and the actual F$_0$ values was less than 1Hz.
Observe that the annealing search program performs significantly
better in all cases.
Note that the mutation operation in the
genetic search program treats each bit in
the parameter encodings equally, while the perturbation
operation in the annealing search program is sensitive
to the distinction between more significant vs.~less
significant bits.  This may explain the better
convergence behaviour of the annealing search.

Notice also in Figure 6 that performance does not degrade
with transcription length as the length doubles from 10 to 20.
This is probably because a randomly generated sequence will
contain downsteps on every second tone (on average)
causing a general downtrend in the F$_0$ values and
severely limiting the combinatorial explosion of possible transcriptions.

\hspace*{-\parindent}
{\bf Trial 2: Artificial Data with Upstep.}
Trial 2 was the same as trial 1 except that this time upstep was
permitted as well.  The results are displayed in Figure 7.
\begin{figure}
\vspace{65mm}
\includegraphics{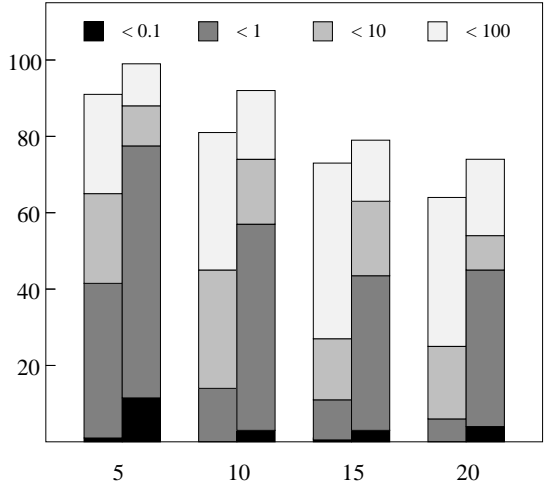}
\caption{Performance results (upstep)}
\end{figure}
Again the annealing program fares better than the genetic program.
Consider again the bars corresponding to evaluations less than 1.
For both programs, however, observe that the performance degrades
more uniformly than in trial 1, probably because the inclusion
of upstep greatly increases the number of possible transcriptions
(and hence, the number of local optima).

\hspace*{-\parindent}
{\bf Trial 3: Actual Data.}
The final trial involved real data, including data from the utterance given
in Figure 1.  This trial involved four subtrials.  The first and second had
F$_0$ sequences of length 10, while the third and fourth had length
18 and 19.  The first and second sequences were taken by extracting
the initial 10 F$_0$ values from the third and fourth sequences,
thereby avoiding the asymptotic behaviour of the longer sequences.
The data is tabulated below, and it
comes from the sentences in (\ref{ex:data}).

\hspace*{-\parindent}
{\small
\begin{tabular}{r|l}
Trial & F$_0$ sequence \\ \hline
1 & 219,168,183,150,160,136,144,123,131,115 \\
2 & 205,224,167,200,156,175,136,156,127,140 \\
3 & 219,168,183,150,160,136,144,123,131,115, \\
  & 122,107,113,105,118,100,113,95\\
4 & 205,224,167,200,156,175,136,156,127,140, \\
  & 118,129,109,119,103,120,102,111,95
\end{tabular}
}

\hspace*{-\parindent}
Performance results are given in Figure 8.
\begin{figure}
\vspace{65mm}
\includegraphics{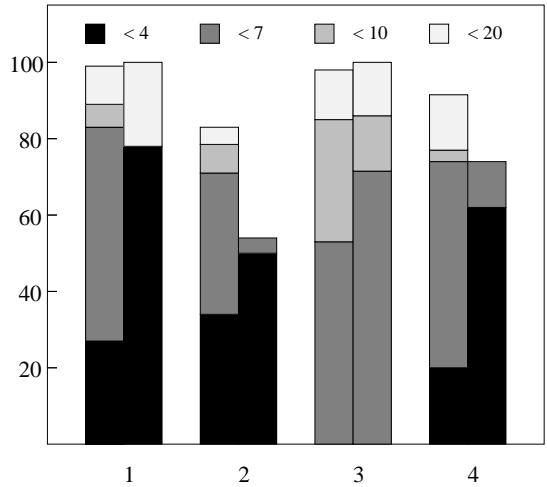}
\caption{Performance results for actual data}
\end{figure}
Notice that the interpretation of the shading in this figure
is different from that in previous figures.  This is because
evaluations near zero were less likely with real data.
In fact, the annealing program never found an evaluation less than 3
while the genetic program never found an evaluation less than 4.

Since the programs performed about equally on finding transcriptions
with an evaluation less than 7, I shall display these transcriptions
along with an indication of how many times each program found the
transcription (G~=~genetic, A~=~annealing).
I give transcriptions which occurred at least twice
in one of the programs, during 100 executions of each.

\hspace*{-\parindent}
{\small
\begin{tabular}{*{3}{@{\hspace{1ex}}l}}
Trial 1: Transcriptions & G & A \\
H\ds L\ds H\ds L\ds H\ds L\ds H\ds L\ds H\ds L &   27 & 37 \\
H\ds L\ds H\ds L\ds HL\ds H\ds L\ds HL &    7 & 0 \\
H\ds L\ds HL\ds HL\ds HL\ds HL &    3 & 0 \\
H\ds LH\ds LH\ds LH\ds LH\ds L &   20 &  2 \\
HL\ds HL\ds HL\ds HL\ds HL &   24 & 39
\end{tabular}
}

\hspace*{-\parindent}
{\small
\begin{tabular}{*{3}{@{\hspace{1ex}}l}}
Trial 2: Transcriptions & G & A \\
L\ds H\ds LH\ds L\ds H\ds L\ds H\ds L\ds H &    5 & 0 \\
L\ds H\ds LH\ds L\ds H\ds LH\ds L\ds H &   66 & 54 \\
\end{tabular}
}

\hspace*{-\parindent}
{\small
\begin{tabular}{*{3}{@{\hspace{1ex}}l}}
Trial 3: Transcriptions & G & A \\
H\ds L\ds H\ds L\ds HL\ds H\ds L\ds HL\ds H\ds L\ds HLH\ds LH\ds L &
  11 & 0 \\
H\ds L\ds HL\ds HL\ds HL\ds HL\ds H\ds L\ds HLH\ds LH\ds L &    1 & 2 \\
H\ds L\ds HL\ds HL\ds HL\ds HL\ds HL\ds HLH\ds LH\ds L &   10 & 14 \\
HL\ds HL\ds HL\ds HL\ds HL\ds HL\ds HLH\ds LH\ds L &   30 & 56 \\
\end{tabular}
}

\hspace*{-\parindent}
{\small
\begin{tabular}{*{3}{@{\hspace{1ex}}l}}
Trial 4: Transcriptions & G & A \\
L\ds H\ds LH\ds L\ds H\ds LH\ds L\ds H\ds L\ds H\ds L\ds H\ds LH\ds
L\ds H\ds L &
60 & 29 \\
L\ds H\ds LH\ds L\ds H\ds LH\ds L\ds HL\ds H\ds L\ds H\ds LH\ds L\ds H\ds L &
5 & 19 \\
L\ds H\ds LH\ds L\ds H\ds LH\ds L\ds HL\ds H\ds L\ds HLH\ds L\ds H\ds L &
7 & 7 \\
L\ds H\ds LH\ds L\ds H\ds LH\ds L\ds H\ds L\ds H\ds L\ds HLH\ds L\ds H\ds L &
0 & 4 \\
L\ds H\ds LH\ds L\ds H\ds LH\ds L\ds HL\ds HL\ds H\ds LH\ds L\ds H\ds L &
0 & 3 \\
L\ds H\ds LH\ds L\ds H\ds LH\ds L\ds HL\ds HL\ds HLH\ds L\ds H\ds L &
0 & 6 \\
\end{tabular}
}

\hspace*{-\parindent}
The results from trial 1 deserve special attention.
In trial 1, three transcriptions were found by both programs.
The best evaluations found are given below:

\hspace*{-\parindent}
{\small
\begin{tabular}{|*{10}{@{\hspace{0.1ex}}r}|*{4}{@{\hspace{1ex}}l}|}
\hline
H&L&\ds H&L&\ds H&L&\ds H&L&\ds H&L&
$\cal E\!: 3$ & $h\!:107$ & $l\!:100$ & $d\!:0.68$ \\
H&\ds L&H&\ds L&H&\ds L&H&\ds L&H&\ds L&
$\cal E\!: 4$ & $h\!:90$ & $l\!:93$ & $d\!:0.76$ \\
H&\ds L&\ds H&\ds L&\ds H&\ds L&\ds H&\ds L&\ds H&\ds L&
$\cal E\!: 3$ & $h\!:107$ & $l\!:100$ & $d\!:0.82$ \\ \hline
\end{tabular}
}

\hspace*{-\parindent}
It is striking to note that the first two transcriptions above
are what Hyman and Stewart (respectively) would have given as transcriptions
for the abstract F$_0$ sequence \mbox{1 3 2 4 3 5 4 6 5 7}.
This is demonstrated in (\ref{equiv}a,b).  The third transcription
points to another possibility, given in (\ref{equiv}c).

\def\yy{$\frac{5}{2}$}
\def\zz#1{$\frac{#1}{2}$}

\begin{ex}
\label{equiv}
\begin{subexamples}
\item {\bf Hyman's transcription scheme} \\
\begin{tabular}{*{10}{@{\hspace{1ex}}c}}
H & L & \downstep H & L & \downstep H & L &
\downstep H & L & \downstep H & L \\ \hline
1 & 3 & 1 & 3 & 1 & 3 & 1 & 3 & 1 & 3 \\
0 & 0 & 1 & 1 & 2 & 2 & 3 & 3 & 4 & 4 \\ \hline
1 & 3 & 2 & 4 & 3 & 5 & 4 & 6 & 5 & 7
\end{tabular}
\item {\bf Stewart's transcription scheme} \\
\begin{tabular}{*{10}{@{\hspace{1ex}}c}}
H & \downstep L & H & \downstep L &
H & \downstep L & H & \downstep L & H & \downstep L \\ \hline
1 & 2 & 1 & 2 & 1 & 2 & 1 & 2 & 1 & 2 \\
0 & 1 & 1 & 2 & 2 & 3 & 3 & 4 & 4 & 5 \\ \hline
1 & 3 & 2 & 4 & 3 & 5 & 4 & 6 & 5 & 7
\end{tabular}
\item {\bf Novel transcription scheme} \\
\begin{tabular}{*{10}{@{\hspace{1ex}}c}}
H & \downstep L & \downstep H & \downstep L & \downstep
H & \downstep L & \downstep H & \downstep L & \downstep
H & \downstep L \\ \hline
1 & \yy & 1 & \yy & 1 & \yy & 1 & \yy & 1 & \yy \\[1mm]
0 & \zz{1} & 1 & \zz{3} & 2 & \zz{5} & 3 & \zz{7} & 4 & \zz{9}\\[1mm] \hline
1 & 3 & 2 & 4 & 3 & 5 & 4 & 6 & 5 & 7
\end{tabular}
\end{subexamples}
\end{ex}

Therefore, there are encouraging signs that the
program is living up to its promise of producing alternative,
equally acceptable transcriptions, as desired from an analytical
standpoint.

\subsection{Multiple Solutions}

Although we have seen more than one transcription for a given
F$_0$ sequence, it is inconvenient to be required to run the programs
several times in order to see if more than one solution can be
found.  Furthermore, the programs are designed not to get caught
in local optima, which is a problem since
interesting alternative transcriptions may
actually be local optima.  Therefore, both programs are set up to
report the $k$ best solutions, where the user specifies the number of
solutions desired.  The program ensures that the same area of
the search space is not re-explored by subsequent searches.
This is done by defining a distance metric on transcriptions
which counts the number of tones in one transcription that have
to be changed in order to make it identical to the other transcription.
That part of the search space within a distance of $n/3$ from any
previously found solution is not explored again.  The programs
give up before finding $k$ solutions if 5
randomly generated transcriptions all fall within
distance $n/3$ of previous solutions.

Now, consider the following randomly generated
sequence of tones:


\hspace*{-\parindent}
\begin{tabular}{|*{7}{@{\hspace{1ex}}l}|l@{\hspace{1ex}}l|}
\hline
\upstep H & \upstep H & \downstep H
& L & \downstep L & \upstep H & L & $h\!:107$ & $l\!:98$ \\
201 & 215 & 201 & 173 & 163 & 201 & 173 & $d\!:0.87$ & $\cal E \!: 0$\\
\hline
\end{tabular}

\hspace*{-\parindent}
The annealing program was set the task of finding ten transcriptions
of this tone sequence.  The program was run only twice, and it reported
the following solutions with evaluations less than or equal to 1.
Both runnings of the program found the same solutions, and in
the same order.  (Note that two transcriptions are taken to be
the same if one or both begin with an initial upstep or downstep;
this has no effect on the phonetic interpretation).
In the following displays, the predicted F$_0$ values are given below
each solution to facilitate comparison with the input sequence.

\hspace*{-\parindent}
\begin{tabular}{|*{7}{@{\hspace{1ex}}l}|l@{\hspace{1ex}}l|}
\hline
\downstep H & \upstep H & \downstep H & L & \downstep L &\upstep H & L &
$h\!:101$ & $l\!:92$ \\
201 & 215 & 201 & 172 & 163 & 201 & 172 & $d\!:0.88$ &
 $\cal E \!: 0.20$ \\ \hline

\downstep H & \upstep H & \downstep H & \upstep L &\downstep L &
 H & \upstep L &
$h\!:109$ & $l\!:94$ \\
201 & 215 & 201 & 174 & 163 & 201 & 174 & $d\!:0.87$ &
$\cal E \!: 0.23$ \\ \hline

L & \downstep H & \downstep H & L & \downstep L &\upstep H & L &
$h\!:105$ & $l\!:97$ \\
201 & 217 & 201 & 174 & 163 & 201 & 174 & $d\!:0.86$ & $\cal E \!: 1.00$ \\
\hline
\end{tabular}
\vspace{2mm}

\hspace*{-\parindent}
\begin{tabular}{|*{7}{@{\hspace{1ex}}l}|l@{\hspace{1ex}}l|}
\hline
H & \upstep H & \downstep H & L & \downstep L &\upstep H & L &
 $h\!:110$ & $l\!:100$ \\
201 & 214 & 201 & 173 & 164 & 201 & 173 & $d\!:0.88$ &
 $\cal E \!: 0.86$ \\ \hline

\upstep H & \upstep H & \downstep H & \upstep L &\downstep L & H &
 \upstep L &
$h\!:102$ & $l\!:88$ \\
201 & 215 & 201 & 174 & 164 & 201 & 174 & $d\!:0.88$ &
$\cal E \!: 0.66$ \\ \hline

\downstep L &\downstep H & \downstep H & L & \downstep L &\upstep H & L &
$h\!:104$ & $l\!:96$ \\
201 & 217 & 201 & 174 & 163 & 201 & 174 & $d\!:0.86$ & $\cal E \!: 1.00$ \\
\hline
\end{tabular}

Since all executions to this point have been based on the first
table of $R$ values, it was decided to try a test with the second
table of $R$ values to see if the performance was different.
Interestingly, the third solution in both of the above executions
was not found, though two new solutions were found.

\hspace*{-\parindent}
\begin{tabular}{|*{7}{@{\hspace{1ex}}l}|l@{\hspace{1ex}}l|}
\hline
\upstep H & \upstep H & \downstep H & L & \downstep L &\upstep H & L &
$h\!:94$ & $l\!:80$ \\
201 & 216 & 201 & 173 & 162 & 201 & 173 & $d\!:0.88$ &
$\cal E \!: 0.49$ \\ \hline

L & \downstep H & \downstep H & L & \downstep L &\upstep H & L &
$h\!:97$ & $l\!:84$ \\
201 & 215 & 201 & 174 & 163 & 201 & 174 & $d\!:0.88$ &
$\cal E \!: 0.65$ \\ \hline

\downstep H & \upstep H & \downstep H & \upstep L &
\downstep L & H& \upstep L &
$h\!:100$ & $l\!:81$ \\
201 & 215 & 201 & 174 & 163 & 201 & 174 & $d\!:0.88$ &
$\cal E \!: 0.92$ \\ \hline

\upstep L & H & L & \downstep H & L & \upstep L &\downstep H &
$h\!:92$ & $l\!:86$ \\
201 & 214 & 201 & 173 & 163 & 201 & 173 & $d\!:0.67$ & $\cal E \!: 0.48$ \\
\hline
\end{tabular}
\vspace{2mm}

\hspace*{-\parindent}
\begin{tabular}{|*{7}{@{\hspace{1ex}}l}|l@{\hspace{1ex}}l|}
\hline
\downstep H & \upstep H & \downstep H & L & \downstep L & \upstep H & L &
$h\!:107$ & $l\!:92$ \\
201 & 216 & 201 & 173 & 162 & 201 & 173 & $d\!:0.87$ &
$\cal E \!: 0.40$ \\ \hline

L & H & L & \downstep H & L & \upstep L & \downstep H &
$h\!:99$ & $l\!:93$ \\
201 & 214 & 201 & 173 & 163 & 201 & 173 & $d\!:0.65$ &
$\cal E \!: 0.82$ \\ \hline

\downstep L & \downstep H & \downstep H & L & \downstep L & \upstep H & L &
$h\!:90$ & $l\!:78$ \\
201 & 217 & 202 & 174 & 163 & 202 & 174 & $d\!:0.88$ & $\cal E \!: 0.86$ \\
\hline
\end{tabular}

Observe that the value of $d$ in the above solutions clusters around
$0.66$ and $0.87$.  Similar clustering may be occurring with the ratio
$h/l$.  However, an analysis of the relationship between the
kinds of solutions found, the two $R$ tables and the parameter
values $h$, $l$ and $d$ has not been attempted.

\subsection{Areas for Further Improvement}

It is rather unsatisfying that the performance of the two
programs is heavily dependent on the setting of several
search parameters, and it seems to be a combinatorial
optimisation problem in itself to find good parameter settings.
My trial-and-error approach will not necessarily have found optimal
parameter values, and so it would be premature to conclude
from the performance comparison that annealing search is
better than genetic search for the problem of tone transcription.
A more thoroughgoing comparison of these two approaches to
the problem needs to be undertaken.

Since the parameters are continuous variables, and since
the evaluation function---which we could write as
${\cal E}_{T,X}(h, l, d)$---is a smoothly
continuous function in $h$, $l$, $d$,
it would be worthwhile to try other (deterministic)
search methods for optimising $h$, $l$ and $d$, once
a candidate tone transcription $T$ has been found.

Finally, it would be interesting to integrate a system like
either of the ones presented here into a speech workstation.
As the phonologist identifies salient points with a cursor the system
would do the transcription, incrementally and interactively.

\section{CONCLUSION}

This paper began with a discussion of the problem of relating tone
transcriptions to their physical counterparts, namely F$_0$ traces.  I
showed that it is desirable for phonologists working on tone to use
sequences of F$_0$ values as their primary data, rather than
impressionistic transcriptions which make (usually implicit)
assumptions about F$_0$ scaling.
I provided an F$_0$ prediction function $\cal P$ which
estimated the F$_0$ value of a tone, given the F$_0$ value of the
previous tone and the identities of the two tones.  I presented
instrumental data from Bamileke Dschang and showed how the
function could be specialised for this language.
The function was then incorporated into the evaluation functions of
two implemented non-deterministic search algorithms.
The performance results were encouraging and
demonstrate the promise of automated tone transcription.

\section*{ACKNOWLEDGEMENTS}

This research is funded by the UK Economic and Social Research
Council, under grant R00023 4439 {\it A Computational Model for the
Phonology-Phonetics Interface in Tone Languages}.  I am indebted to
SIL Cameroon for their logistical support on my field trip
in September and October of 1993,
during which the data presented in the paper (and much other data
besides) was gathered, and especially to
Nancy Haynes, Gretchen Harro for helping me collect the
data and Jean-Claude Gnintedem who endured many recording
sessions.
I am grateful to John Coleman, Michael Gasser and Marie South
for helpful comments on an earlier version of this paper.
The F$_0$ data was extracted using the ESPS Waves+
package in the Edinburgh University Phonetics Laboratory.

\end{document}